\documentclass[journal]{IEEEtran}
\usepackage{amsmath,amsfonts}
\usepackage{algpseudocode}
\usepackage{algorithm} 
\usepackage{array}
\usepackage[caption=false,font=footnotesize,labelfont=rm,textfont=rm]{subfig}
\usepackage{textcomp}
\usepackage{stfloats}
\usepackage{url}
\usepackage{verbatim}
\usepackage{graphicx}
\usepackage{amsmath}
\usepackage{xcolor}
\usepackage[colorlinks,linkcolor=blue,anchorcolor=blue,citecolor=blue]{hyperref}
\usepackage{cite} 
\usepackage{setspace}
\newcommand{\orcid}[1]{\href{https://orcid.org/#1}{\includegraphics[width=10pt]{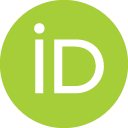}}}
\usepackage{makecell}
\usepackage{multirow} 
\begin{document}

\title{Attention-Guided Sequence Feature Fusion and Mixed-Space Metric Learning for Cross-View Image Sequence Geo-Localization}

\author{
\IEEEauthorblockN{
Nan~Wu, 
Chunfang~Yang, 
Jiangshan~Li,
Xiangyang~Luo
\thanks{This work was supported in part by the National Natural Science Foundation of China (No.U23A20305, No.62472440), and in part by the Innovation Scientists and Technicians Troop Construction Projects of Henan Province (No.254000510007).}
\thanks{Nan~Wu, Chunfang~Yang, Xiangyang~Luo are with Information Engineering University, Zhengzhou, 450001, China, and the Henan Key Laboratory of Cyberspace Situation Awareness, Zhengzhou, 450001, China. (e-mail: myhofzy@163.com; chunfangyang@126.com; luoxy\_ieu@sina.com). \textit{(Corresponding author: Chunfang Yang.)} }
\thanks{Jiangshan~Li is with the College of Electronic Science and Technology, National University of Defense Technology, Changsha 410073, China. (e-mail:lijiangshan@nudt.edu.cn)} } }



\maketitle

\begin{abstract}
Cross-view image sequence geo-localization aims to determine the geographic location of a sequence of ground images by matching it against geo-tagged satellite imagery. Although sequential observations provide complementary scene information, substantial overlap between adjacent images introduces feature redundancy, while conventional fusion strategies may weaken spatial relationships essential for cross-view matching. Therefore, we propose a cross-view image sequence geo-localization method via \textbf{A}ttention-guided \textbf{S}equence fusion and \textbf{M}etric \textbf{L}earning (ASML). Specifically, a Sequential Content Fusion (SCF) module models cross-image feature correlations to suppress redundant content and adaptively preserve geographically informative non-overlapping regions. Next, a Spatial Feature Aggregation (SFA) module is used to capture and aggregate the spatial and semantic information of local features, so as to maintain the spatial layout of the image sequence. Finally, a Geo-InfoNCE loss function is proposed, which comprehensively considers the feature distance and geographic distance for mixed-space metric learning, thus improving the ability of the model to distinguish hard negative samples. A new cross-view image sequence dataset CrossRoad-66K is established to provide more complex scenarios. Extensive experiments and visualizations on the SeqGeo and CrossRoad-66K datasets demonstrate that compared to state-of-the-art (SOTA) methods, the proposed method ASML achieves an improvement of 3.37\% and 2.10\% at R@1, respectively.
\end{abstract}


\section{Introduction}
\IEEEPARstart{c}{ross-view} image sequence geo-localization is a technique that retrieves the most similar image from a database of geo-tagged aerial images (usually satellite images) for a ground image sequence, thus estimating the geographic location where the ground image was captured \cite{ref2,ref62,ref63,ref58,ref64}. Previous studies focused on matching a single ground image with aerial images in a reference dataset \cite{ref51,ref53,ref56}. In recent years, with the rise of short-form video social media platforms, numerous ground image sequences have been produced quickly. These ground image sequences contain richer semantic information than a single image. Therefore, it is significantly meaningful to geo-localize the ground image sequence in many applications, such as autonomous driving\cite{ref5}, augmented reality\cite{ref52}, drone navigation\cite{ref32}.
\begin{figure}[!t]
\centering
\subfloat[Average ratio of overlapping area under different sequence lengths]{
\label{fig1_1}
\includegraphics[width=3.1 in]{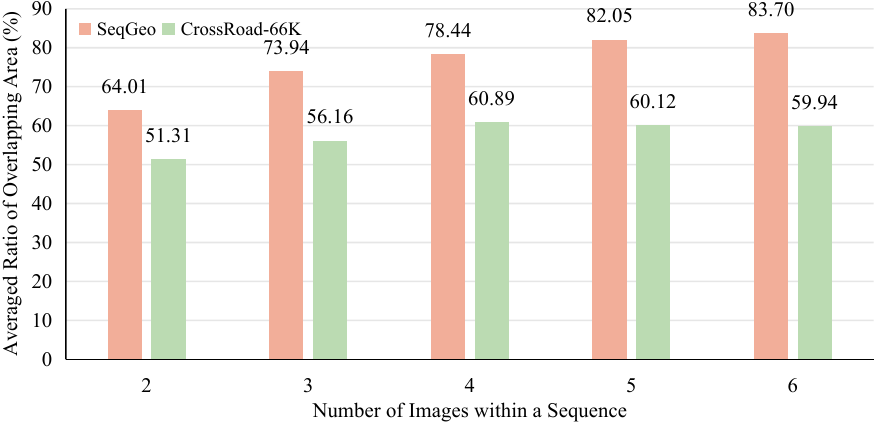}}\\
\subfloat[Visualization of the overlapping areas in ground images]{
\label{fig1_2}
\includegraphics[width=3.1 in]{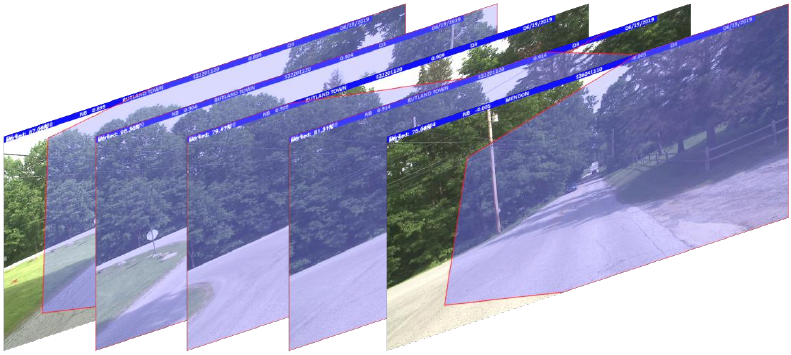}}
\caption{Quantitative and qualitative analysis of overlapping areas in cross-view image sequences. (a) The average ratio of overlapping areas among ground image sequences with different sequence lengths on the SeqGeo and CrossRoad-66K datasets. (b) Visualization of overlapping regions between adjacent ground images. This figure provides a dataset-level analysis of sequence redundancy rather than a performance comparison between different methods.}
\label{fig1}
\end{figure}

Cross-view image sequence geo-localization is an extension of cross-view image geo-localization. Existing cross-view image geo-localization methods \cite{ref9,ref11,ref38,ref49,ref65,ref66} aim to minimize the distance between the features of the matched ground image and satellite image while maximizing the distance between the features of the unmatched ones. Cross-view image sequence geo-localization methods \cite{ref23, ref25, ref24, ref43} essentially follow the same principles. The difference lies on extracting the feature representation of the image sequence by fusing the local features of multiple ground images in the sequence, which is significant for subsequent feature matching.

The early cross-view image sequence methods extract the feature of each ground image independently, and then concatenate these features along the sequence dimension \cite{ref24}. Recently, researchers \cite{ref43} propose to select the important components from the features of ground images by multi-head attention mechanism. \textbf{However, these methods do not consider the influence of overlapping areas in the image sequence, which may introduce redundant information and disrupt the spatial layout of objects in different images.} Fig.~\ref{fig1} analyzes the inherent redundancy characteristics of image sequences in different datasets, which motivates the design of the proposed overlap-aware feature selection strategy. In other words, if the redundant content from overlapping areas are not effectively suppressed, it would weaken the discriminative ability of the feature representation and decrease the geo-localization accuracy.

Therefore, this work proposes a cross-view image sequence geo-localization method via attention-guided sequence feature fusion and metric learning (ASML). In the proposed method, the Sequential Content Fusion (SCF) module is designed to fuse geographic-related features from sequential ground images while eliminating redundant content through the attention-guided feature selection. The Geo-InfoNCE loss function is presented to perform mixed-space metric learning, which simultaneously combines feature space and geographic distance to guide model optimization. Extensive experiments on the SeqGeo and CrossRoad-66K datasets demonstrate that ASML significantly improves geo-localization accuracy and reduces computational costs compared to existing methods.

The main contributions can be summarized as follows.

\begin{itemize}
    \item 
    We propose a cross-view image sequence geo-localization method ASML, with a task-specific Sequential Content Fusion (SCF) module. Unlike conventional attention-based feature fusion strategies, SCF transforms cross-image correlation into an overlap-aware selection mechanism, enabling redundancy suppression and spatially ordered feature reconstruction.
    \item
    A Geo-InfoNCE loss function is designed to integrate feature space and geographic space. It enables more accurate quantification of the prediction error by a mixed-space alignment mechanism and guides the optimization of the model.
    \item
    We establish a new cross-view image sequence dataset CrossRoad-66K to provide a complementary benchmark with urban-dominant scenarios. Compared with existing datasets, CrossRoad-66K contains more complex scenarios, enabling a more comprehensive evaluation of cross-view image sequence geo-localization methods. 
    \item
    Extensive experiments demonstrate that, compared to existing methods, the proposed ASML achieves improvements of 3.37\% and 2.10\% at R@1 on the SeqGeo and CrossRoad-66K datasets, respectively, while effectively controlling computational costs.
\end{itemize}

The remainder of this manuscript is organized as follows. Section~\ref{sec2} reviews the related work on cross-view image geo-localization and image sequence geo-localization. Section~\ref{sec3} discusses the key challenge of cross-view image sequence geo-localization that this work aims to address. Section~\ref{sec4} provides a detailed introduction to the proposed method ASML and its core modules. Section~\ref{sec5} presents extensive experiments and analyses of ASML, and compares its performance with existing methods. Finally, Section~\ref{sec6} summarizes the main contributions of this work.

\section{Related Work}\label{sec2}
In this section, we briefly review existing works on cross-view image geo-localization and image sequence geo-localization.
\subsection{Cross-View Image Geo-Localization}
The key points of cross-view image geo-localization lie in the feature aggregation approach and the loss function. 

Due to the drastic difference in spatial layouts between ground and satellite views, researchers developed a series of feature aggregation methods to enhance the representational capability of ground image and satellite image features. For example, Hu \textit{et al.}\cite{ref12} proposed CVM-Net, which added a NetVLAD\cite{ref13} layer on the VGG16 network to embed ground images and satellite images in a unified feature space. Shi \textit{et al.}\cite{ref14} added two consecutive fully connected layers after the backbone network as the Spatial Aware Feature Aggregation Module (SAFA), which encodes the scenery information in ground and satellite images to improve the robustness of feature representation. To reduce the view difference between ground images and satellite images, Ye \textit{et al.} \cite{ref48} transformed street-view panoramas into BEV (Bird's Eye View) image and combine panorama-to-satellite and BEV-to-satellite retrieval branches to improve geo-localization accuracy. Ouyang \textit{et al.} \cite{ref49} proposed CLGT, which integrates causal learning and BEV-based geometric topology fusion to improve the robustness and discriminability of cross-view image features.

The loss function also plays a significant role in enhancing the model's generalization ability and robustness. In the early years, Hu \textit{et al.}\cite{ref12} proposed a weighted soft-margin triplet loss, which adds a scaling factor to amplify the loss based on the soft-margin triplet loss. To further improve the model's ability to distinguish hard negative samples, Cai \textit{et al.} \cite{ref17} and Guo \textit{et al.} \cite{ref18} introduced a hard sample mining strategy that differentiates the simple and hard samples within a batch and re-weights them.

Existing methods have achieved impressive performance, some of them have been extended to cross-view image sequence geo-localization. However, because these methods independently extract features from each ground image in an image sequence, it is difficult for them to capture the spatial layout of the objects in the realistic scenario, leading to poor geo-localization performance.

\subsection{Image Sequence Geo-Localization}
Image sequence geo-localization can be divided into same-view and cross-view image sequence geo-localization according to the view of the reference image. In the early years, same-view image sequence geo-localization is the mainstream methods. For example, Gonzalo \textit{et al.}\cite{ref19} segmented street view videos into query frames, then matched the handcrafted features of query frames and reference ground images to estimate the trajectory of the query video. Regmi \textit{et al.}\cite{ref21} designed a temporal feature extraction network to extract the features of query frames and reference ground images and proposed a deep learning approach to correct the offsets of estimated GPS positions.

In recent years, with the improvement of satellite image resolution, cross-view image sequence geo-localization has attracted increasing attention. Vyas \textit{et al.}\cite{ref23} matched video-frame features with aerial image patches to localize the geographic region covered by a video. Shi \textit{et al.}\cite{ref59} proposed CVLNet, which projects sequential ground-view features into the overhead view and aggregates them into a global sequence representation while estimating relative displacement for cross-view alignment. Pillai \textit{et al.}\cite{ref25} further adopted a coarse-to-fine strategy, first retrieving a large aerial region using the global video feature and then matching individual frames with local aerial patches for trajectory prediction.

For sequence-level representation learning, Zhang \textit{et al.}\cite{ref24} randomly masked ground-image features and concatenated the retained features to suppress redundant content. Yuan \textit{et al.}\cite{ref43} employed multi-head attention between adjacent ground images to aggregate temporal features. More recently, Wu \textit{et al.}\cite{ref50} introduced the Set-CVGL task and the FlexGeo framework to adaptively fuse features from unordered ground-view image sets for cross-view geo-localization.

However, existing methods mainly focus on sequence-level feature aggregation or coarse image-level selection, while the redundant overlapping content between adjacent ground images is not explicitly modeled.

\begin{figure*}[!t]
\centering
\includegraphics[width=0.8\linewidth]{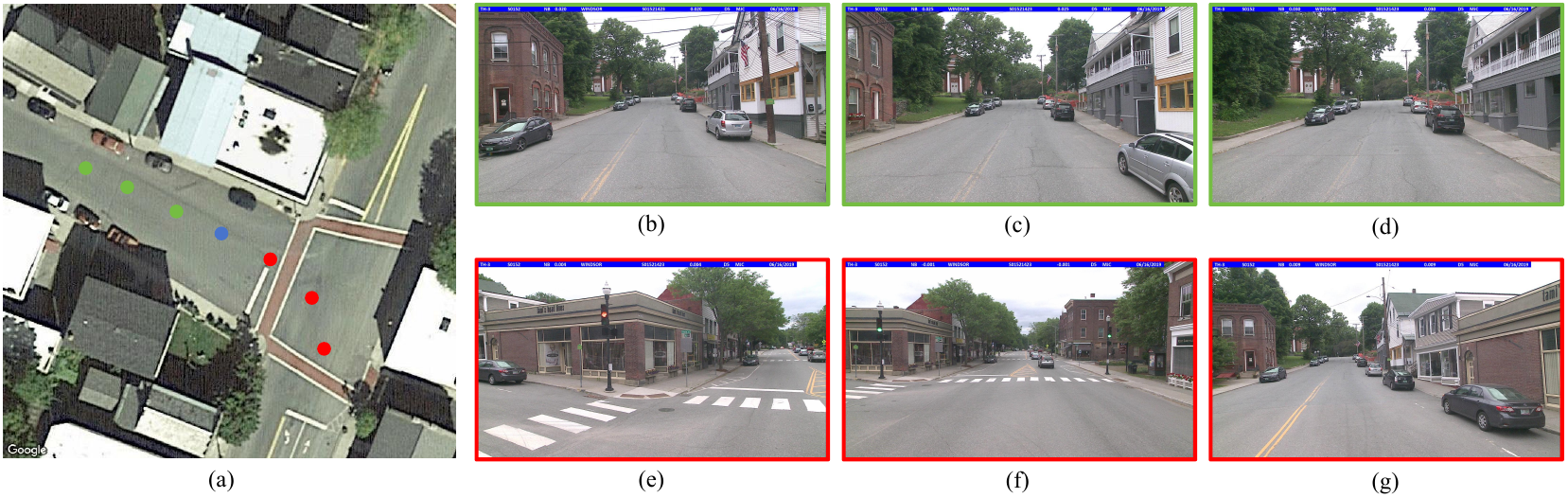}
\caption{A pair of samples in SeqGeo dataset. (a) shows the satellite image. (b)-(d) show the ground images captured along a straight road, whose locations are marked by green points in (a). (e)-(g) show the ground images captured around a corner, whose locations are marked by red points in (a).}
\label{fig3}
\end{figure*}
\section{Problem Statement}\label{sec3}

Given a ground image sequence $S^{\mathrm{grd}}=\{I_i^{\mathrm{grd}}\}_{i=1}^{K}$, where $K$ is the sequence length, cross-view image sequence geo-localization aims to retrieve the satellite image covering the geographic area captured by the sequence from a reference gallery $\{I_j^{\mathrm{sat}}\}_{j=1}^{R}$. In this task, multiple ground images are first encoded and compressed into a sequence-level descriptor:
\begin{equation}
d^{\mathrm{grd}}=\mathcal{A}\left(f(I_1^{\mathrm{grd}}),f(I_2^{\mathrm{grd}}),\cdots,f(I_K^{\mathrm{grd}})\right),
\end{equation}
where $f(\cdot)$ denotes the image feature extractor and $\mathcal{A}(\cdot)$ denotes the sequence-level feature fusion function. The geo-localization result is then obtained by matching $d^{\mathrm{grd}}$ with the descriptor of each reference satellite image $d_j^{\mathrm{sat}}$:
\begin{equation}
b=\underset{j \in {1,\cdots,R}}{\operatorname{arg\min}}~
D\left(d^{\mathrm{grd}}, d_j^{\mathrm{sat}}\right),
\end{equation}
where $D(\cdot,\cdot)$ denotes the feature distance, and $b$ represents the index of the matched satellite image.

As shown in Fig.~\ref{fig1_2}, adjacent ground images usually contain large overlapping regions, where scenery objects such as buildings, roadsides, trees, and traffic facilities repeatedly appear. To evaluate the influence of such redundancy, we randomly select 100 sample pairs from SeqGeo and compare the similarity between the fused ground image sequence feature and the corresponding satellite feature under different random masking settings. The similarity is measured by L2 distance, cosine similarity, G2S KL divergence, and S2G KL divergence. In Table~\ref{tab0}, Max Masks denotes the maximum number of ground images randomly removed from an input sequence in SeqGeo~\cite{ref24}. Larger values of Max Masks indicate that more images may be discarded before feature concatenation, thus reducing part of the redundant content.

\begin{table}
\caption{Comparative Analysis of Ground Image Sequence Feature with Satellite Image Feature Using Structural Similarity and KL Divergence. $\downarrow$ Indicates Lower Values with Greater Similarity and $\uparrow$ Indicates Higher Values with Greater Similarity.}
\label{tab0}
\centering
\resizebox{0.98\linewidth}{!}{
\renewcommand{\arraystretch}{1.1}
\begin{tabular}{ccccc}
\Xhline{1.3pt}
\multirow{2}{*}{Max Masks} & \multicolumn{2}{c}{Structural Similarity}  & \multicolumn{2}{c}{KL Divergence}\\
\cline{2-5}
~ &  L2 Distance$\downarrow$ & Cosine Similarity$\uparrow$ & G2S($10^{-5}$)$\downarrow$  & S2G($10^{-5}$)$\downarrow$ \\ 
\Xhline{0.8pt}
        5 & 0.1033 & 0.0465 & 1.7548 & 1.7551\\
        3 & \textbf{0.1019} & \textbf{0.0480} & \textbf{1.6820} & \textbf{1.6826}\\
        1 & 0.1021 & 0.0478 & 1.6935 & 1.6945\\
        0 & 0.1024 & 0.0475 & 1.7083 & 1.7088\\
\Xhline{1.3pt}
\end{tabular}
}
\end{table}

As shown in Table~\ref{tab0}, these results suggest that both excessive redundancy and overly aggressive random masking may weaken cross-view feature similarity. Therefore, coarse image-level masking is insufficient, demanding local overlap-aware feature selection.

Spatial layout preservation is another key issue. For straight-road sequences, adjacent ground images often maintain horizontal continuity, so direct feature concatenation may partially preserve object layout. However, for corner-road sequences, abrupt trajectory changes break this continuity, and direct concatenation or random masking may distort the spatial relationship among geographic objects, especially when the ground observations correspond to a turning road structure in the satellite view.

Therefore, the main challenge is not merely aggregating multiple ground image features, but constructing an overlap-aware sequence representation that suppresses redundant overlapping contents while preserving the spatial organization of non-overlapping geographic-related features for cross-view matching.

\begin{figure*}[t]
\centering
\includegraphics[width=0.98\linewidth]{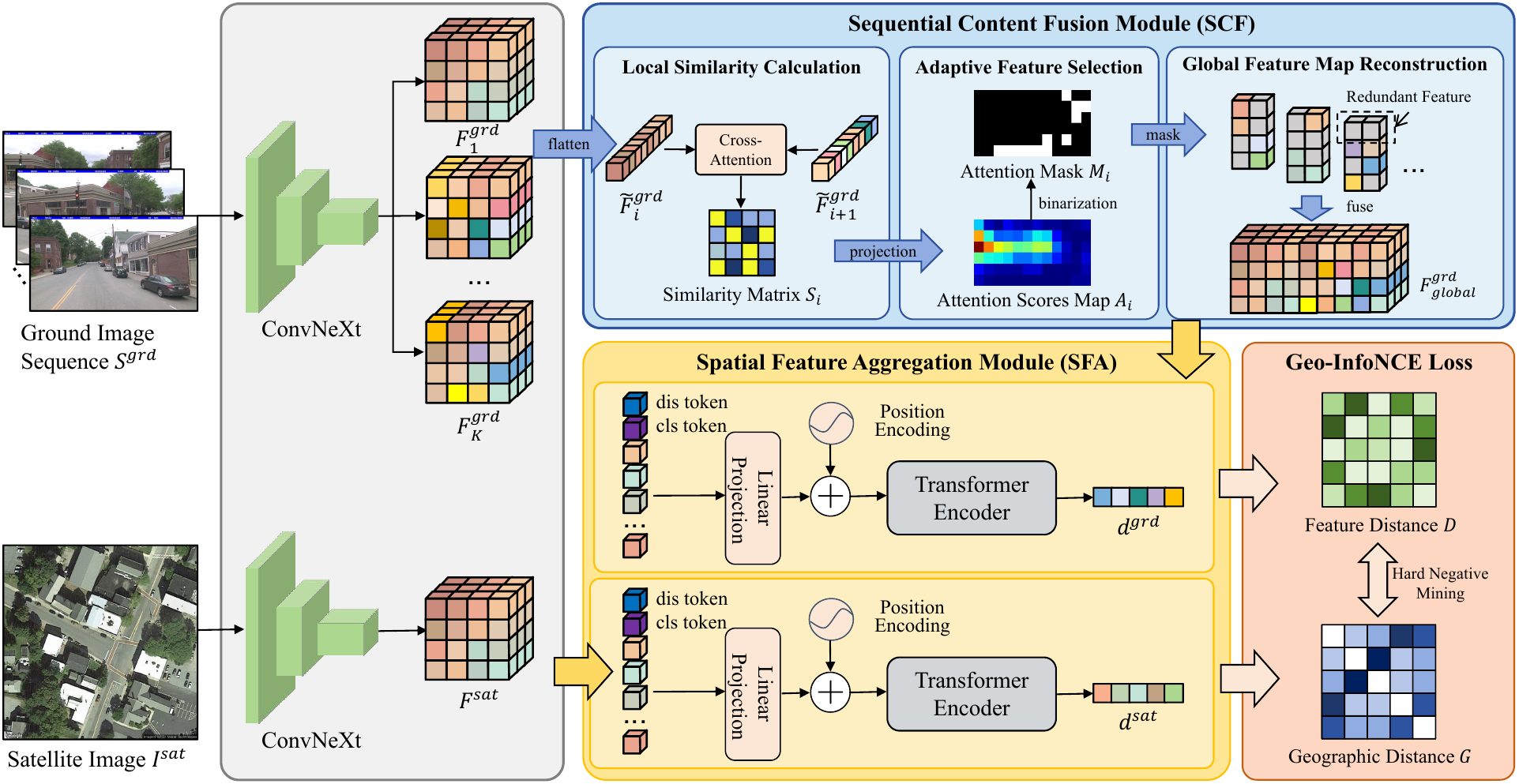}
\caption{Overview architecture of ASML. Specifically, the Sequential Content Fusion Module (SCF) integrates the geographical semantics among the ground image sequence, and reduces the number of tokens in the subsequent processing. Spatial Feature Aggregation Module (SFA) mainly contains a Transformer Encoder, which is used to aggregate feature descriptors. The Geo-InfoNCE Loss is used to learn discriminative features in feature space and geographic space.}
\label{fig4}
\end{figure*}
\section{Methodology}\label{sec4}
Inspired by the analysis in Section \ref{sec3}, the most intuitive idea is to eliminate redundant information while maintaining the spatial layout information of the objects within the ground images. Therefore, we propose a cross-view image sequence geo-localization method via attention-guided sequence feature fusion and metric learning (ASML). In this section, we introduce the architecture and each component of ASML in detail.

\subsection{Method Overview: ASML}\label{sec4_1}
As shown in Fig.~\ref{fig4}, the proposed method ASML mainly consists of the following components: (1) a ConvNeXt backbone \cite{ref30} for feature extraction, (2) a Sequential Content Fusion (SCF) module for global feature integration, (3) a Spatial Feature Aggregation (SFA) module for feature aggregation, and (4) a Geo-InfoNCE loss function for mixed-space metric learning.

These components constitute two pipelines: the ground image pipeline and the satellite image pipeline. The ground image pipeline takes the ground image sequence $S^{\mathrm{grd}}$ in an image sequence as input. First, we feed them into the ConvNeXt backbone to extract local semantic features. The extracted feature maps $F_{i}^{\mathrm{grd}}$ are then inputted into the SCF module to obtain the global ground feature map $F_{\mathrm{global}}^{\mathrm{grd}}$. Next, the global feature map is aggregated into a ground image descriptor $d^{\mathrm{grd}}$ through the SFA module. The satellite image pipeline takes the satellite image $I^{\mathrm{sat}}$ as input. Similarly, the ConvNeXt backbone extracts the local feature map $F^{\mathrm{sat}}$ from the satellite image, which is then aggregated into the satellite image descriptor $d^{\mathrm{sat}}$ using the SFA module. Finally, the Geo-InfoNCE loss function is used to measure the difference between the ground image feature descriptor $d^{\mathrm{grd}}$ and the satellite image feature descriptor $d^{\mathrm{sat}}$, allowing for optimization of model parameters.

Notably, ASML constructs the image sequence descriptor through global feature reconstruction and descriptor aggregation, while the Transformer-based SFA module further aggregates the cross-view feature map into a compact descriptor for matching. Therefore, SFA serves as a descriptor aggregation component rather than a generic sequence fusion module.

\begin{figure*}[t]
\centering
\includegraphics[width=0.9\linewidth]{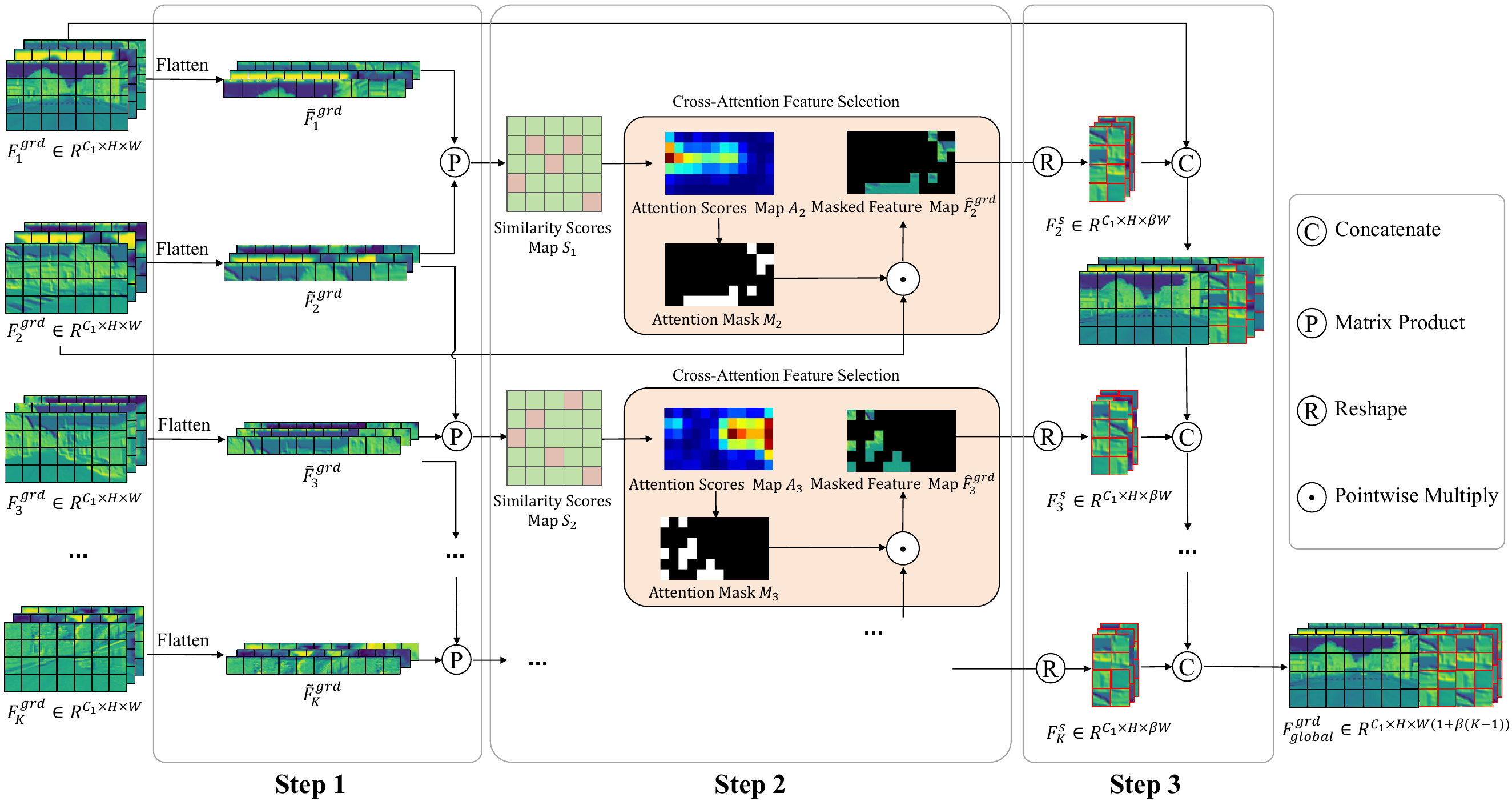}
\caption{The operational workflow of the Sequential Content Fusion Module (SCF). First, a similarity scores map $S_{i}$ is calculated between the adjacent local features. Next, an attention scores map $A_{i+1}$ is generated from similarity scores map, which is then translated to an attention mask $M_{i+1}$. Finally, the attention mask is utilized to select the features in non-overlapping areas and eliminate redundant features.}
\label{fig5}
\end{figure*}
\subsection{Sequential Content Fusion Module (SCF)}\label{sec4_2}
Unlike direct concatenation or generic Transformer-based fusion, the proposed Sequential Content Fusion (SCF) module performs task-specific feature selection before aggregation, as shown in Fig.~\ref{fig5}. It calculates the local correlations between adjacent ground images to identify the overlapping regions, then preserves the non-overlapping regions that provide richer geographic information. The selected features are further reconstructed into a spatially ordered global feature map. 

Different from conventional cross-attention modules that aim to enhance feature interaction, SCF uses correlation scores as an indicator of spatial redundancy rather than feature importance. Specifically, SCF implements the sequence-level feature fusion through three steps: local similarity calculation, adaptive feature selection, and global feature map reconstruction.

\subsubsection{Step 1: Local Similarity Calculation} The inputs of the SCF module are $K$ sequential local feature maps $F_{i}^{\mathrm{grd}}\in \mathbb{R}^{C_1\times H\times W}$, where $C_1, H,$ and $W$ represent the channel dimension, height, and width of the local ground feature map, respectively. Through reshaping operations along the spatial dimension, the three-dimensional feature maps are flattened to the two-dimensional version $\tilde{F}_{i}^{\mathrm{grd}}\in \mathbb{R}^{C_1\times (H\times W)}$. 
Then, the local similarity matrix $S_{i}\in \mathbb{R}^{(H\times W)\times (H\times W)}$ is calculated between the adjacent feature maps.
\begin{equation}
S_{i}=(\tilde{F}_{i}^{\mathrm{grd}})^{\mathrm{T}} \times {\tilde{F}_{i+1}^{\mathrm{grd}}} ,
\end{equation}
where $S_{i}(m,n),\;m,n \in[1, H \times W]$ represents the dot product between the $m$-th local feature of $\tilde{F}_{i}^{\mathrm{grd}}$ and the $n$-th local feature of $\tilde{F}_{i+1}^{\mathrm{grd}}$. Low similarity scores indicate weak attention correlation between the paired areas, which geometrically corresponds to the non-overlapping areas between adjacent ground images. Therefore, the objective of SCF is opposite to conventional attention weighting strategies: instead of emphasizing highly correlated regions, it suppresses redundant correlated regions and retains complementary geographic information.

\subsubsection{Step 2: Adaptive Feature Selection}
To identify potentially non-overlapping areas, we first generate an attention mask $M_{i+1}\in\{0,1\}^{H\times W}$ through:
\begin{equation}
M_{i+1}(k,l)=
\left\{
\begin{array}{ll}
1, & \text{if } A_{i+1}(k,l)<\tau_i(k),\\
0, & \text{otherwise},
\end{array}
\right.
\end{equation}
\begin{equation}
A_{i+1}(k,l)=
\max_{1\leq j\leq H\times W}
S_i\big(j,(k-1)W+l\big),
\end{equation}
\begin{equation}
\tau_i(k)=
\text{Quantile}\left(A_{i+1}(k,:),\beta_i\right),
\end{equation}
where $A_{i+1}$ is the attention scores map of $F_{i+1}^{\mathrm{grd}}$, and $(k-1)W+l$ denotes the flattened spatial index corresponding to position $(k,l)$. $\tau_i(k)$ is the row-wise quantile threshold of the attention scores, and $\beta_i\in[\beta_{\min},\beta_{\max}]$ denotes the adaptive feature selection ratio for the $i$-th pair of adjacent ground images. The selection ratio $\beta_i$ is adaptively determined from the attention-score distribution. Accordingly, the lowest-scoring features in each row are selected to suppress redundant content.

The generated attention mask $M_{i+1}$ is a binary matrix, whose one-valued elements indicate regions with relatively low cross-image correlation, which are potentially non-overlapping areas in the feature map $F_{i+1}^{\mathrm{grd}}$. Then, the attention mask $M_{i+1}\in\{0,1\}^{H\times W}$ is broadcast to $M'_{i+1}\in\{0,1\}^{C_1\times H\times W}$ by replicating it $C_1$ times along the channel dimension, which is used to extract the selected local features:
\begin{equation}
\hat{F}_{i+1}^{\mathrm{grd}}
=F_{i+1}^{\mathrm{grd}}\odot M'_{i+1},
\end{equation}
where $\odot$ denotes point-wise multiplication, and $\hat{F}_{i+1}^{\mathrm{grd}}$ is the masked feature map.

\subsubsection{Step 3: Global Feature Map Reconstruction}
First, we reshape $\hat{F}_{i+1}^{\mathrm{grd}}$ to select its geographic-related features and adjust its spatial dimension. Since the adaptive selection ratio $\beta_i$ may vary across adjacent image pairs, a fixed output width $W_{\max}=\lceil\beta_{\max}W\rceil$ is adopted to ensure consistent feature dimensions:
\begin{equation}
F_{i+1}^{s}
=\text{Padding}
\left(
\text{Reshape}
\left(\hat{F}_{i+1}^{\mathrm{grd}}\right),{W_{\max}}
\right),
\label{e7}
\end{equation}
where the Reshape operation compacts the selected features in each row according to their original spatial order, and the Padding operation fills the remaining positions with zeros to maintain a fixed output width. Specifically, approximately $\lceil\beta_i W\rceil$ features are retained per row, while the output feature map has a fixed dimension $F_{i+1}^{s}\in \mathbb{R}^{C_1\times H\times W_{\max}}$. This process preserves the relative spatial order of the selected features and constructs a compact representation for subsequent descriptor aggregation.

By repeating the above operations on subsequent images in the sequence, $K-1$ feature maps $F_{i+1}^{s}$, $i\in\{1,\ldots,K-1\}$, are obtained. Finally, a global feature map $F_{\mathrm{global}}^{\mathrm{grd}} \in\mathbb{R}^{C_1\times H\times [W+(K-1)W_{\max}]}$ is reconstructed as follows:
\begin{equation}
F_{\mathrm{global}}^{\mathrm{grd}}
=F_1^{\mathrm{grd}}
\Vert F_2^s
\Vert\cdots\Vert F_K^s,
\end{equation}
where $\Vert$ denotes width-wise concatenation.

In summary, the SCF module adaptively fuses sequential local feature into a global feature map through attention-guided spatial feature selection, thus preserving spatial layout and eliminating redundant content. Unlike generic feature fusion, SCF is specifically designed for the redundancy and spatial layout challenges in cross-view image sequence geo-localization. As no additional learnable parameters are introduced, it can also control computational cost.
\begin{figure}[!t]
\centering
\subfloat[]{
\label{fig6_1}
\includegraphics[width=0.45\linewidth]{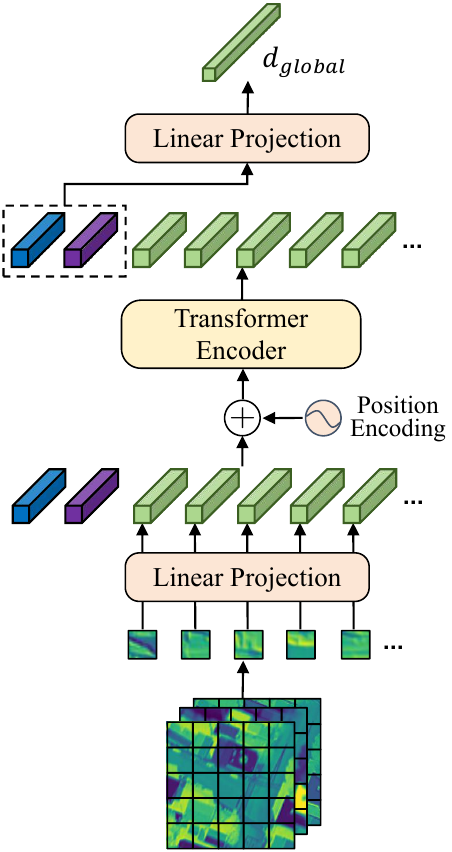}}
\quad
\subfloat[]{
\label{fig6_2}
\includegraphics[width=0.4\linewidth]{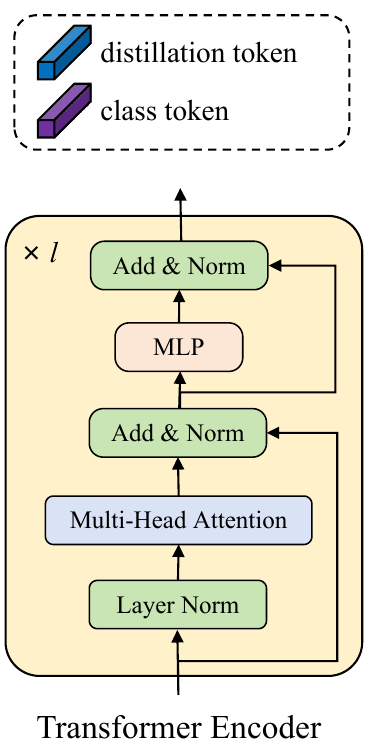}}
\caption{The main processing of the spatial feature aggregation module (SFA). (a) shows the workflow of SFA module. (b) shows the detailed structure of the Transformer Encoder in (a).}
\label{fig6}
\end{figure}
\subsection{Spatial Feature Aggregation Module (SFA)}\label{sec4_3}
Feature aggregation modules \cite{ref12,ref13,ref14} are commonly used to enhance spatial and semantic consistency between matched ground-satellite features. Following recent works \cite{ref25,ref28,ref46}, we adopt DeiT \cite{ref31} as the Spatial Feature Aggregation Module (SFA), since its class and distillation tokens can aggregate local features and capture positional-semantic correlations to produce discriminative global descriptors.

The specific processing of the SFA module is illustrated in Fig.~\ref{fig6}. The input feature map is first decomposed into $N$ unit-scale (1×1) patches. These patches are then fed into a linear projection layer to generate the corresponding patch tokens $t_i\in \mathbb{R}^{C_2}, i \in [1,N]$, where $C_2$ is the channel dimension of each token. Then, the distillation token $t_{\mathrm{dis}}$ and the class token $t_{\mathrm{cls}}$ are inserted into patch tokens, where $t_{\mathrm{dis}}$ serves as an auxiliary learnable aggregation token to provide complementary discriminative information and $t_{\mathrm{cls}}$ is used to collect global contextual information from all patch tokens. These tokens are augmented with a learnable positional encoding $E_{\mathrm{pos}}\in \mathbb{R}^{C_2\times (N+2)}$:
\begin{equation}
X_0 = [t_{\mathrm{dis}},t_{\mathrm{cls}},t_1,t_2,\cdots,t_{N}] + E_{\mathrm{pos}},
\end{equation}
where $X_0$ is the input of the Transformer Encoder. The positional encoding $E_{\mathrm{pos}}$ corresponds to the positions in the reconstructed global feature map, rather than the original coordinates of the unmasked feature map. Therefore, it can model the relative and global spatial order of the compact sequence-level representation. Finally, the global feature descriptor $d_{\mathrm{global}}$ is generated as follows:
\begin{equation}
X_l = T(X_0) = [d_{\mathrm{dis}},d_{\mathrm{cls}}, d_1, d_2, \cdots,d_{N}],
\end{equation}
\begin{equation}
d_{\mathrm{global}} = \operatorname{LP}(d_{\mathrm{dis}})+\operatorname{LP}(d_{\mathrm{cls}}),
\end{equation}
where $T$ represents the $l$-layer Transformer Encoder, $X_l$ is the output of the $l$-layer Transformer Encoder, $\operatorname{LP}$ represents linear projection, $d_i\in \mathbb{R}^{C_2}, i \in [1,N]$ are the output of patch tokens. 
Notably, when the input is the global ground feature map $F^{\mathrm{grd}}_{\mathrm{global}}$, $d_{\mathrm{global}}$ corresponds to the ground descriptor $d^{\mathrm{grd}}$. When the input is the satellite feature map $F^{\mathrm{sat}}$, $d_{\mathrm{global}}$ corresponds to the satellite descriptor $d^{\mathrm{sat}}$. 
Relying on the powerful multi-head attention mechanism of DeiT, the distillation token $d_{\mathrm{dis}}$ and the class token $d_{\mathrm{cls}}$ can effectively integrate the global contexts of each token, thus keeping the spatial layout information.

In summary, the SFA module is adopted as a descriptor aggregation component to encode the reconstructed global feature map. It can capture the spatial layout of scenery objects in ground images and aggregate semantically discriminative global feature descriptors, thus significantly enhancing cross-view image geo-localization robustness.
\begin{figure}[!t]
\centering
\subfloat[]{\label{fig8_1}\includegraphics[width=0.4\linewidth]{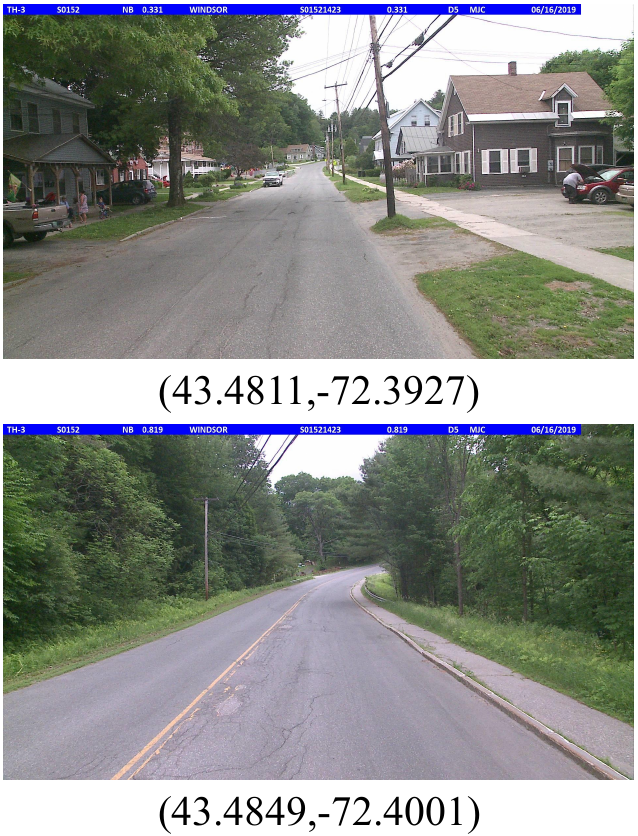}}
\quad
\subfloat[]{\label{fig8_2}\includegraphics[width=0.23\linewidth]{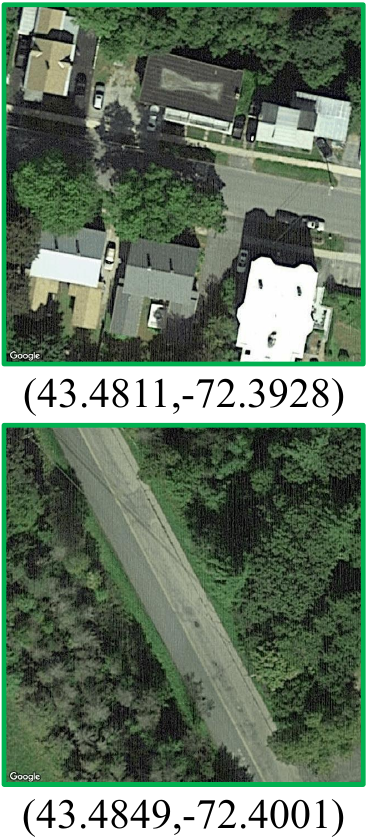}}
\quad
\subfloat[]{\label{fig8_3}\includegraphics[width=0.23\linewidth]{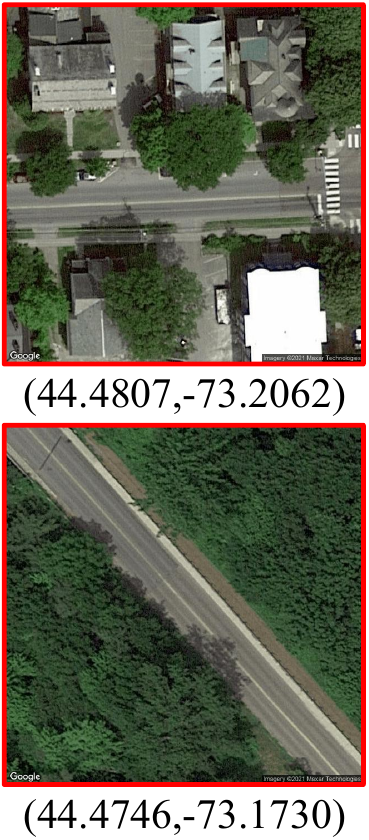}}
\caption{Visualization of some hard negative samples in the SeqGeo dataset, which are easily confused. (a) shows one image within the ground image sequence. (b) shows the matched satellite image. (c) shows the unmatched satellite image. The left ground image and the right satellite image are a pair of difficult samples. Each image is labeled with the GPS coordinates of the capture location.}
\label{fig8}
\end{figure}
\begin{figure*}[t]
\centering
\includegraphics[width=0.8\linewidth]{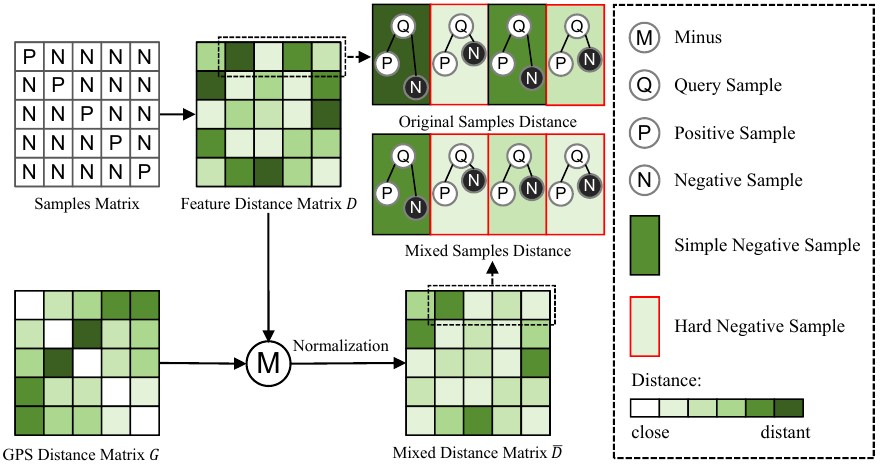}
\caption{The mixed distance metric optimization strategy for the samples within a batch. The samples matrix represents the relationship between the samples within a batch, where the elements on the diagonal are positive samples, while the others are negative samples. $M$ refers to the feature distance matrix $D$ minus the GPS distance matrix $G$.}
\label{fig7}
\end{figure*}

\subsection{Geo-InfoNCE Loss Function}\label{sec4_4}

Cross-view image geo-localization aims to pull matched ground-satellite pairs closer in the feature space while pushing unmatched pairs apart. The InfoNCE loss\cite{ref33} has been widely adopted in contrastive representation learning and cross-view geo-localization tasks~\cite{ref29,ref40}. It can be formulated as:
\begin{equation}
\mathcal{L}_{\mathrm{InfoNCE}}=-\log \frac{\exp\left((1-D_{i, i}) / \tau\right)}{\sum_{j = 1}^{B} \exp\left((1-D_{i, j})/ \tau\right)},
\end{equation}
\begin{equation}
D_{i, j} =1/2 \left\lVert \mathbf{q}_i-\mathbf{k}_j \right\rVert_2=1- \mathbf{q}_i \cdot \mathbf{k}_j,
\end{equation}
where $\left\lVert \mathbf{q}_i-\mathbf{k}_j \right\rVert_2$ denotes the L2 distance between the $i$-th query image's feature vector $\mathbf{q}_i$ and the $j$-th reference image's feature vector $\mathbf{k}_j$, $i,j \in [1,B]$, $B$ is the batch size, and $\tau$ is a learnable temperature parameter. Although InfoNCE effectively optimizes feature-space similarity, it does not explicitly consider the geographic distance between samples. Therefore, visually similar but geographically distant satellite images may be treated as hard negative samples and interfere with model optimization, as shown in Fig.~\ref{fig8}. Existing works have introduced geographic constraints into the loss function \cite{ref34,ref45,ref57}, but they are usually formulated as an additional matching constraint and may still provide insufficient penalties for visually similar but geographically distant hard negatives.

To address this issue, we propose a Geo-InfoNCE loss function $\mathcal{L}_{\mathrm{Geo}}$, which integrates feature distance and geographic distance into a unified mixed-space metric:
\begin{equation}
\mathcal{L}_{\mathrm{Geo}}=-\log \frac{\exp\left((1-\overline{D}_{i, i}) / \tau\right)}{\sum_{j = 1}^{B} \exp\left((1-\overline{D}_{i, j})/ \tau\right)},
\end{equation}
\begin{equation}
\overline{D}_{i, j}=D_{i, j}-\lambda_g \cdot \mathrm{Norm}(G_{i, j}),
\end{equation}
\begin{equation}
G_{i, j}=\sqrt{\left(\mathrm{lon}_{i}-\mathrm{lon}_{j}\right)^{2}+\left(\mathrm{lat}_{i}-\mathrm{lat}_{j}\right)^{2}},
\end{equation}
where $\overline{D}_{i,j}$ denotes the mixed distance between the $i$-th query feature and the $j$-th reference feature. $D_{i,j}$ represents the feature distance between the corresponding feature representations. $G_{i,j}$ denotes the GPS distance between the center points of the satellite images corresponding to the $i$-th and $j$-th samples within a batch. $\lambda_g$ is a scaling coefficient that controls the contribution of geographic distance in the mixed-space metric. $\mathrm{Norm}(\cdot)$ denotes a batch-wise normalization operation applied to the geographic distance matrix. Therefore, Geo-InfoNCE serves as a task-oriented mixed-space metric for contrastive optimization, where \(D_{i,j}\) measures the discrepancy between the learned visual representations, while \(G_{i,j}\) provides complementary geographic information to suppress visually similar but geographically inconsistent negative pairs.

It should be noted that $D$, $G$, and $\overline{D}$ are all $B \times B$ matrices, whose elements are scalar terms rather than high-dimensional vectors. Empirically, the feature distance $D$ and the geographic distance $G$ in city-scale datasets exhibit comparable numerical magnitudes and remain within the same order of magnitude. Therefore, this lightweight batch-wise normalization is sufficient to stabilize their numerical scales without introducing additional learnable parameters or a more complex mapping function. The scaling coefficient $\lambda_g$ is further introduced to adjust the relative contribution of geographic distance, and its effectiveness is verified by an ablation study. Since the samples are captured from the same city, the distance deviation caused by the Earth's curvature has negligible influence on the loss optimization. Therefore, we calculate the GPS distance using the L2 distance through latitude and longitude.

Generally, Geo-InfoNCE loss $\mathcal{L}_{\mathrm{Geo}}$ integrates feature space and geographic space through a mixed-space alignment mechanism, as demonstrated in Fig.~\ref{fig7}. It can prevent the model from the interference caused by hard negative samples, thus better guiding the optimization of the model parameters.
\section{Experiment}\label{sec5}
\subsection{Cross-View Image Sequence Datasets}\label{sec5_1}
The experiments were performed on SeqGeo and CrossRoad-66K datasets, which are introduced as follows.
\subsubsection{SeqGeo Dataset}
SeqGeo dataset\cite{ref24} is a large-scale cross-view image sequence dataset collected in Vermont, USA, comprising 38,863 ground image sequences and 38,863 satellite images. This dataset was split into a training set with 31,091 image pairs and a testing set with 7,772 image pairs. The ground images in this dataset were collected on approximately 500 $km$ of urban and suburban roads, with around 70\% from suburban areas and 30\% from urban areas. Each ground image sequence contains 7 images, with 8 meters between the capture coordinates of adjacent images, and the FoV (field-of-view) of each ground image is approximately 120\({}^{\circ}\).
\begin{table*}
\caption{Comparison between the proposed CrossRoad-66K dataset and existing cross-view image datasets.}
\label{tab2}
\centering
\resizebox{0.85\linewidth}{!}{
\renewcommand{\arraystretch}{1.1}
\begin{tabular}{c|ccccc}
\Xhline{1.3pt}
Datasets & CVACT\cite{ref39}  & CVUSA\cite{reference2} & VIGOR\cite{ref26}  & SeqGeo\cite{ref24} & CrossRoad-66K\\ 
\Xhline{0.8pt}
        Satellite Images & 44,416 & 44,416 & 90,618 & 38,863 & 21,320\\
        Ground Images & 44,416 & 44,416 & 105,214 & 118,549 & 66,384\\
        Ground Images per Satellite Image  & 1 & 1 & 5 & 7 & 6\\
        Seamless Sampling & $\times$& $\times$& $\checkmark$& $\checkmark$& $\checkmark$\\
        Sequential Ground Images & $\times$& $\times$& $\times$& $\checkmark$& $\checkmark$\\
        Full Panorama & $\checkmark$& $\checkmark$& $\checkmark$& $\times$& $\times$\\
        Ground Image Center Orientation & North & North & North & Random & Random\\
        Coverage & Urban, Suburban & Urban, Suburban & Suburban & Urban, Suburban & Urban, Suburban\\
\Xhline{1.3pt}
\end{tabular} 
}
\end{table*}
\begin{table}
\caption{Hyperparameter tuning for the quantile threshold $\beta$.}
\label{tab3}
\centering
\resizebox{0.95\linewidth}{!}{
\begin{tabular}{ccccccc}
\Xhline{1.5pt}
\multirow{2}{*}{$\beta$}&
         \multicolumn{3}{c}{SeqGeo} & \multicolumn{3}{c}{CrossRoad-66K} \\ 
         \cline{2-7}
       ~ & R@1 & R@5 & R@1\%  &  R@1 & R@5 & R@1\%      \\ 
\Xhline{0.8pt}               
         8.3\%  & 	15.89\%  & 	34.61\%  & 	74.15\%  & 	4.57\%  & 	13.02\%  &	38.30\%   \\
        16.7\%  &	21.53\%  &	41.46\%   &	79.04\%  &	5.25\% &	15.75\% &	42.82\%   \\
       25\%  &	21.34\%  &	41.69\%  &	78.44\% &	4.99\%  &	14.81\%   &	41.76\%  \\
       33.3\%  &	19.97\%  &	38.45\%   &	76.37\%  &	4.61\%  &	14.24\%   &	41.38\% \\
       Adap.  &	\textbf{23.53\%} &	\textbf{44.30\%} &	\textbf{80.77\%}  &	\textbf{6.20\%}  &	\textbf{16.83\%}  & \textbf{43.94\%}  \\
\Xhline{1.5pt}
\end{tabular} 
}
\end{table}

\subsubsection{CrossRoad-66K Dataset}
We establish a new cross-view image sequence dataset CrossRoad-66K to evaluate the performance of the proposed method in challenging scenarios. It is an urban-dominant dataset that contains more complex scenarios such as road structures, denser buildings, and diverse variations. The specific establishment process of the CrossRoad-66K dataset is as follows.

\textbf{Ground Image Sampling:} Real driving video data was collected on approximately 400 $km$ of roads in Ziyang City, Sichuan Province, China, combined with the corresponding GPS coordinates. Then these driving videos were segmented into frames as ground images, with a size of 2560 × 1180 and a FoV of approximately \({140}^{\circ}\). The total number of ground images is 66,384, and the distance between the capture coordinates of adjacent images is 8 meters. Similar to the SeqGeo dataset, every 6 successive ground images are grouped into a ground image sequence. 
Due to the consecutive sampling strategy of the satellite images in the CrossRoad-66K dataset, there are overlapping areas between adjacent satellite images, which results in some ground images being covered by two or three satellite images. Therefore, the total number of ground images is not six times of the satellite images.

\textbf{Satellite Image Sampling:} Since each ground image sequence contains 6 images, the average GPS coordinate of all ground images in a ground image sequence was calculated as the central coordinate. The satellite image at the central coordinate of each ground image sequence was then downloaded from Bing Maps. In order to cover all the capture coordinates of the images in each ground image sequence, the satellite images are 300 × 300 pixels in size and have a spatial resolution of 0.298 $\mathrm{m/pixel}$.

The CrossRoad-66K dataset was split into a training set containing 17,056 image pairs and a testing set containing 4,264 image pairs. The training and testing sets were collected from different regions. The differences between the CrossRoad-66K dataset and existing cross-view image datasets are shown in Table~\ref{tab2}.

\subsection{Implementation Details and Evaluation Metrics}\label{sec5_2}
\textbf{Implementation Details:} ASML was implemented in PyTorch and deployed on two NVIDIA RTX 3090 GPUs (24GB). ConvNeXtV2-F pre-trained on ImageNet-1k was utilized as the backbone. DeiT-S pre-trained on ImageNet-21k was utilized as the SFA module, which was the same as TransGeo\cite{ref28}. The number of layers $l$ in DeiT-S was set to 12 while the patch size was set to 1. Each experiment was trained for 200 epochs with a batch size of 64 using the AdamW optimizer with an initial learning rate of 0.0001, and the weight decay coefficient was set to 0.03. The ground image size and the satellite image size were set to 192 × 384 and 320 × 320, respectively. The scaling coefficient $\lambda_g$ of Geo-InfoNCE Loss is set to 0.75. To ensure the fairness of experiments, all the experimental results used for comparison are obtained from their papers or evaluated using their public code. 
\begin{table*}
\caption{Comparison with existing methods. \textbf{boldfaced} texts denote the best results and \underline{underlined} texts denote the suboptimal results.}
\label{tab4}
\centering
\resizebox{0.85\linewidth}{!}{
\begin{tabular}{ccccccccc}
\Xhline{1.5pt}
\multirow{2}{*}{Methods}&
         \multicolumn{4}{c}{SeqGeo} & \multicolumn{4}{c}{CrossRoad-66K} \\ 
         \cline{2-9}
       ~ & R@1 & R@5 & R@10 & R@1\%   & R@1 & R@5 & R@10 & R@1\%     \\ 
\Xhline{0.8pt}     
        SeqGeo\cite{ref24} &2.07\% & 8.12\% & 13.16\% & 40.10\% & 1.86\%	& 7.69\%	& 13.56\%	& 35.65\%\\
        GeoDTR+\cite{ref37}	& 2.52\%	& 7.29\%	& 11.69\%	& 36.85\%	& 1.29\%	& 3.89\%	& 6.32\%	& 17.93\%\\
        GeoDTR\cite{ref36}	& 3.20\%	& 9.03\%	& 14.08\%	& 39.98\%	& 3.15\%	& 9.14\%	& 14.18\%	& 27.33\%\\
        GAReT\cite{ref25} & 3.34\% & 11.19\% & 17.18\% & 44.39\% & \underline{4.10\%} & \underline{14.24\%} & \underline{21.46\%} & 40.48\% \\
        L2LTR\cite{ref27} & 3.78\% & 12.00\% & 18.21\% & 47.59\% & 2.53\% &	9.71\% &	15.74\% &	23.15\%\\
        TransGeo\cite{ref28} & 9.38\% & 22.43\% & 29.81\% & 57.41\% & 2.37\%	& 9.85\%	& 15.62\%	& 34.47\%\\
        Sample4Geo\cite{ref29} & \underline{20.16\%} & \underline{39.63\%} & \underline{49.38\%} & \underline{75.24\%} & 2.35\%	& 7.34\%	& 11.37\%	& 23.57\%\\
        FlexGeo\cite{ref50} & 3.51\% & 14.22\% & 20.77\% & 48.53\% & 3.82\% & 14.17\% & 21.15\% & \underline{41.41\%} \\
        \textbf{ASML (Ours)}	& \textbf{23.53\%} &	\textbf{44.30\%} &	\textbf{53.23\%} & \textbf{80.77\%}	& \textbf{6.20\%}  &	\textbf{16.83\%} &	\textbf{24.40\%} & \textbf{43.94\%} \\
\Xhline{1.5pt}
\end{tabular} 
}
\end{table*}

The feature selection ratio $\beta_i$ in SCF controls the proportion of low-correlation features retained from each ground image. A small $\beta_i$ may discard informative geographic features, whereas a large $\beta_i$ may introduce redundant content. To accommodate varying overlap characteristics across image sequences, we adaptively determine $\beta_i$ from the attention-score distribution, with $\beta_{\min}=8.3\%$ and $\beta_{\max}=33.3\%$. We further compare the adaptive strategy (Adap.) with different fixed selection ratios, as reported in Table~\ref{tab3}. The adaptive strategy consistently outperforms all evaluated fixed ratios across the retrieval metrics on both SeqGeo and CrossRoad-66K, demonstrating the effectiveness of dynamically adjusting the feature selection budget according to cross-image correlations.

\begin{figure*}[ht]
\centering
\includegraphics[width=0.7\linewidth]{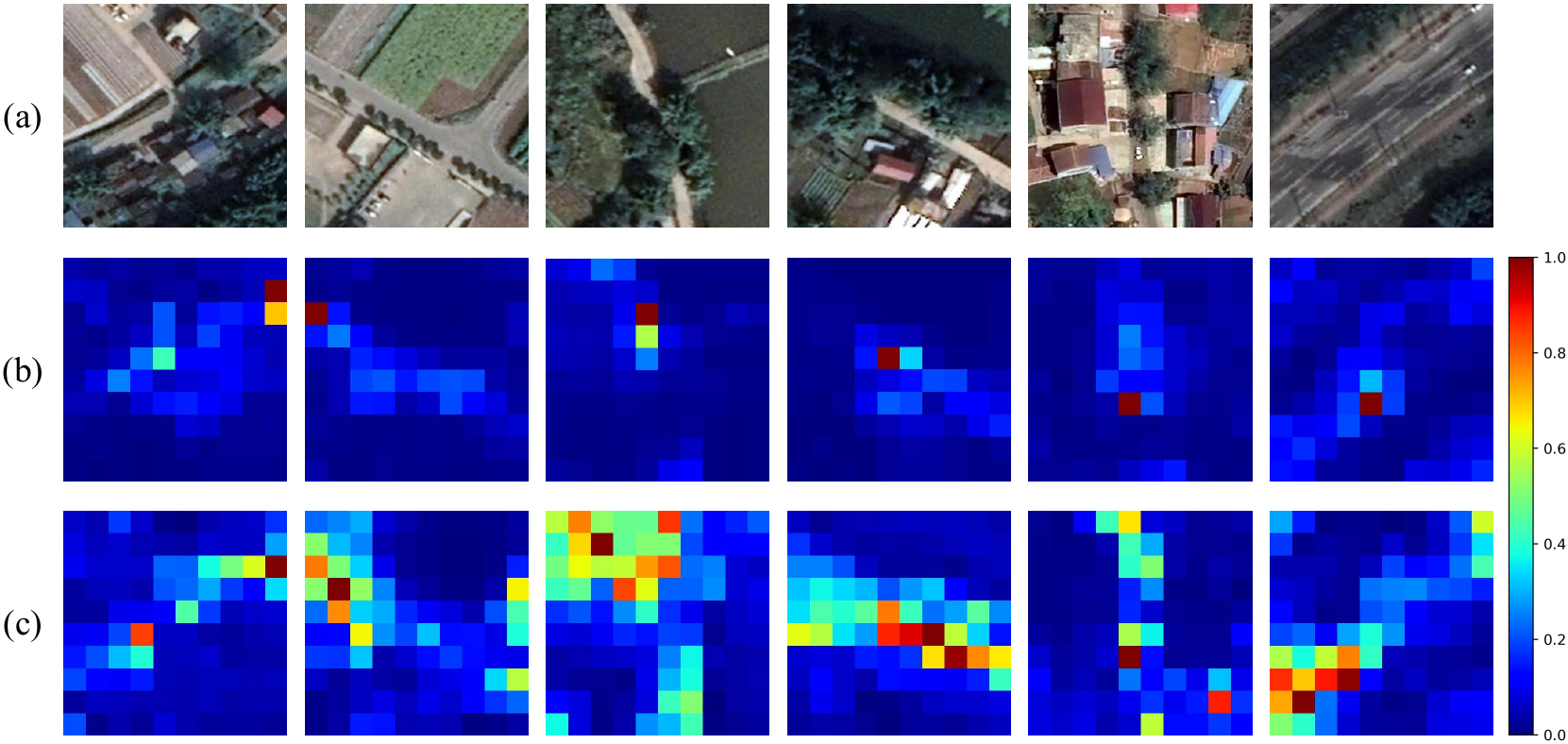}
\caption{Visualization comparison of attention maps between the baseline method and ASML. (a) shows the satellite images, (b) presents the attention maps generated by the baseline method\cite{ref28}, and (c) shows the attention maps produced by the proposed ASML. Compared with the baseline, ASML generates more concentrated responses on geographically discriminative regions while suppressing redundant regions, demonstrating its effectiveness in selecting informative features for cross-view matching.}
\label{fig10}
\end{figure*}

\textbf{Evaluation Metrics.} We adopt the recall rate at top-$k$ (R@$k$) and the recall rate at 1\% (R@1\%) to measure the retrieval accuracy. Given a query ground image sequence, R@$k$ and R@1\% represent the probability of the correct satellite image ranking within the top-$k$ and top-1\% retrieval results, respectively. In the experiments, we evaluate the performance of all methods using the following four evaluation metrics: R@1, R@5, R@10, and R@1\%.
\begin{figure*}[ht]
\centering
\subfloat[Ground Images]{
\label{fig14_1}
\includegraphics[width=0.125\linewidth]{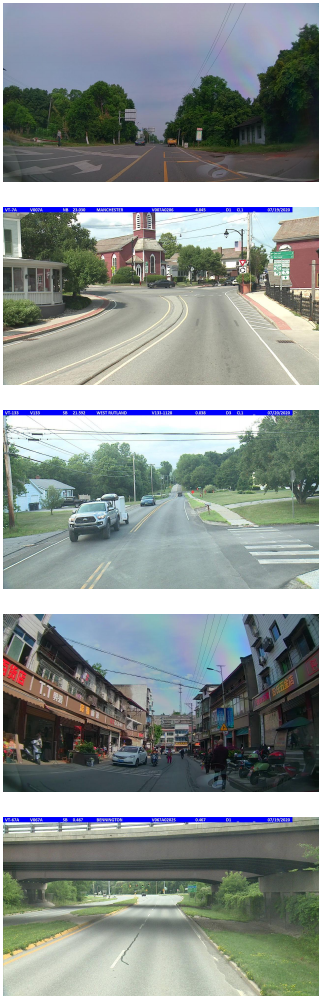}}
\quad
\subfloat[Ground-Truth]{
\label{fig14_2}
\includegraphics[width=0.0735\linewidth]{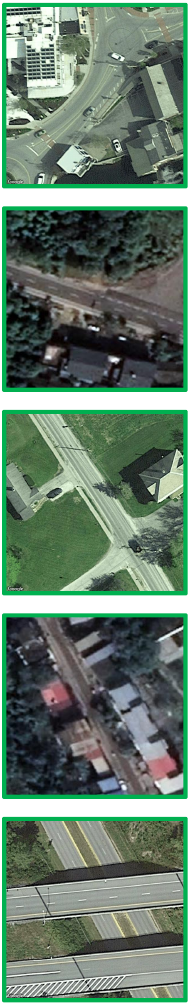}}
\quad
\subfloat[Top-3 Results from ASML]{
\label{fig14_3}
\includegraphics[width=0.225\linewidth]{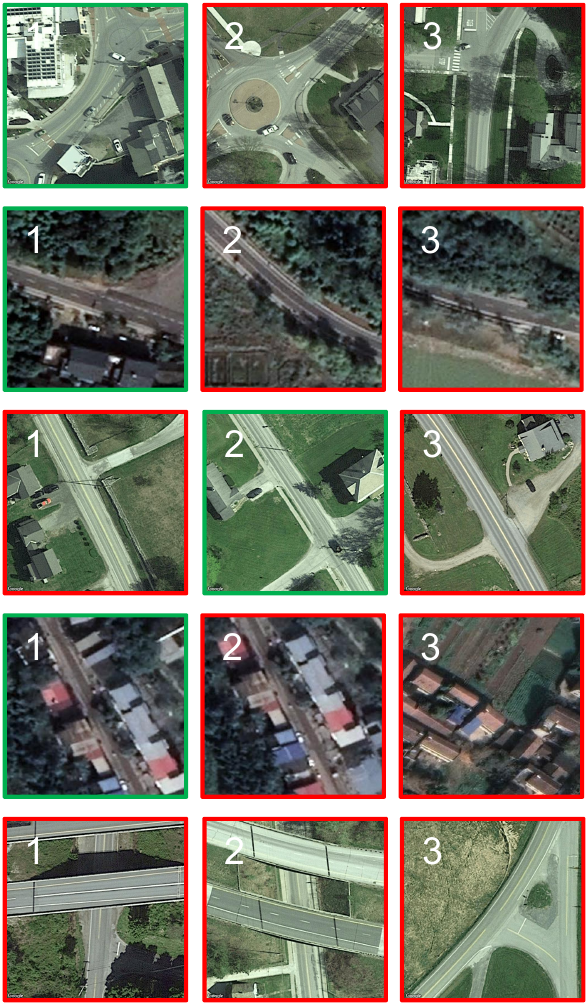}}
\quad
\subfloat[Top-3 Results from Sample4Geo \cite{ref29}]{
\label{fig14_4}
\includegraphics[width=0.225\linewidth]{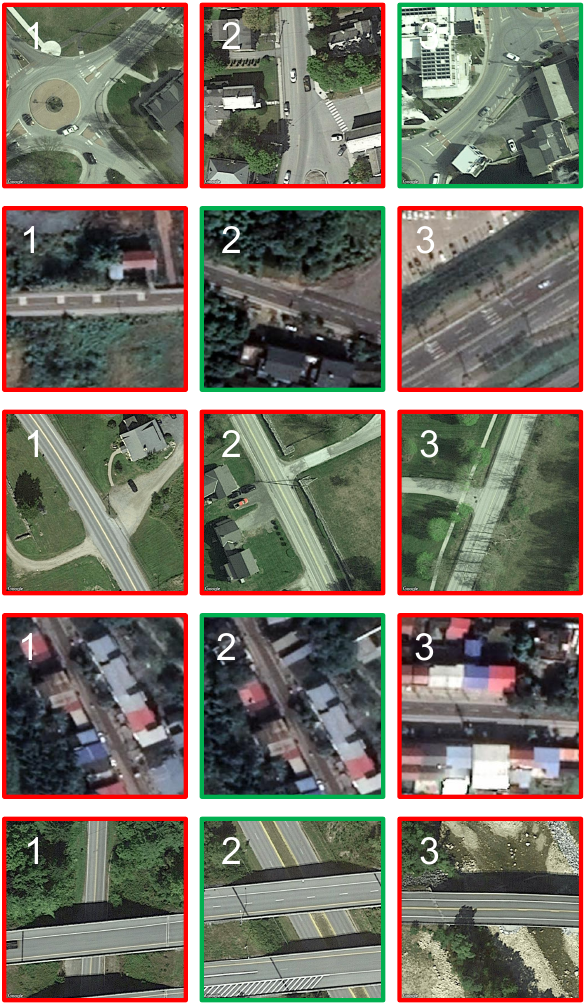}}
\quad
\subfloat[Top-3 Results from GAReT \cite{ref25}]{
\label{fig14_5}
\includegraphics[width=0.225\linewidth]{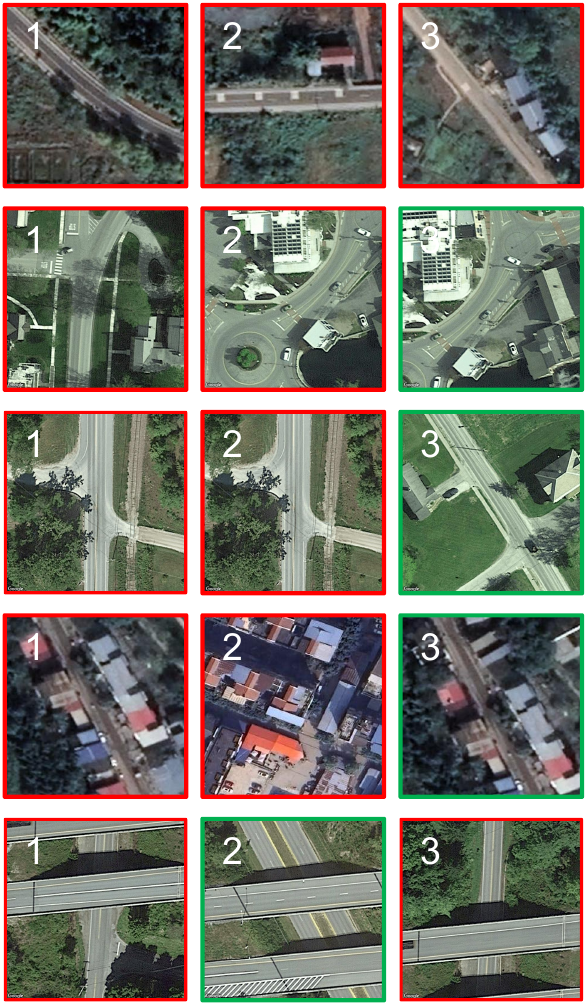}}
\caption{Qualitative comparison of ASML and other methods. (a) shows the first images of the query ground image sequences. (b) shows the ground-truth results. (c) shows the Top-3 satellite images returned from ASML. (d) shows the Top-3 satellite images returned from Sample4Geo. (e) shows the Top-3 satellite images returned from GAReT. The first and second rows show the geo-localization results of dense-building areas, the third and fourth rows show the results of sparse-building areas, and the last row shows the results of overpass-intersection areas. The satellite images marked with green boxes denote the ground-truth results, which provide a common reference for all compared methods. The images marked with red boxes represent incorrect retrieval results. The first and fourth rows are samples from the CrossRoad-66K dataset, while the rest are samples from the SeqGeo.}
\label{fig14}
\end{figure*}
\subsection{Comparing ASML with State-of-the-art Methods}\label{sec5_3}
\subsubsection{Quantitative Performance Comparisons}
The performance of the proposed ASML and several advanced methods was tested on the SeqGeo and CrossRoad-66K datasets. For single-image methods\cite{ref27,ref28,ref29,ref36,ref37}	, we adapted them to the sequence setting by concatenating each ground image feature. For sequence-based methods, we followed their original sequence processing strategy. Table~\ref{tab4} shows the R@1, R@5, R@10, and R@1\% of these methods on the two datasets. On the SeqGeo dataset, ASML outperforms the suboptimal method Sample4Geo \cite{ref29} with 3.37\%, 4.67\%, 3.85\%, and 5.53\% at R@1-R@1\%, respectively. On the more challenging CrossRoad-66K dataset, ASML outperforms the suboptimal methods with 2.10\%, 2.59\%, 2.94\%, and 2.53\% at R@1-R@1\%. It demonstrates that explicit geographic feature selection and redundant information removal are important for image sequence geo-localization. It can also be observed that there is a significant gap in the recall rates across the two datasets, indicating that the complex scenario of the CrossRoad-66K dataset is a real challenge for geo-localization.

\subsubsection{Attention Map Analysis}

As shown in Fig.~\ref{fig10}, the baseline method tends to assign attention responses to broad visual regions, including redundant areas that are less beneficial for cross-view localization. In contrast, the proposed ASML produces more focused attention distributions on geographically informative structures, such as distinctive building layouts and road configurations. This difference originates from the Sequential Content Fusion (SCF) module, which explicitly models the correlation between adjacent ground images and suppresses redundant overlapping information before feature aggregation.

\subsubsection{Geo-Localization Results Visualization}
The geo-localization results obtained from ASML and other methods are visualized in Fig.~\ref{fig14}. It shows several geo-localization results in three typical scenarios: dense-building areas, sparse-building areas, and overpass-intersection areas.

\textbf{Geo-Localization Results in Dense-Building Areas:} The first and second rows of Fig.~\ref{fig14} show the geo-localization results in dense-building areas. For the given ground images, ASML can match the target satellite image accurately, while Sample4Geo fails to do so. This is because the images captured in dense-building areas contain more scenery objects, whose spatial layout is relatively complex. The proposed ASML not only eliminates redundant content, but also keeps spatial layout information. In contrast, when Sample4Geo is used for image sequence geo-localization, the redundant content among the ground images cannot be removed. Therefore, it will cause repeated appearance of certain scenery features and the loss of spatial layout information, thus adversely affecting geo-localization performance. 

\textbf{Geo-Localization Results in Sparse-Building Areas:} The third and fourth rows of Fig.~\ref{fig14} show the geo-localization results in sparse-building areas. For these given images, ASML also achieves better geo-localization performance than Sample4Geo. This is because the images captured in sparse-building areas contain fewer scenery objects, whose spatial layout is relatively simple, resulting in more hard negative samples in sparse-building areas. Therefore, the Geo-InfoNCE loss function can guide ASML to better distinguish these hard negative samples during training. Consequently, ASML exhibits great geo-localization performance even in sparse-building areas.  

\textbf{Geo-Localization Results in Overpass-Intersection Areas:} The last row of Fig.~\ref{fig14} shows the geo-localization result in overpass-intersection area. For the given ground image, neither ASML nor Sample4Geo is able to geo-localize the correct satellite image. Occlusions at overpasses and intersections can cause certain scene elements to be visible in ground images but absent from satellite imagery. It would introduce interference to the geo-localization task.

\begin{table*}
\caption{Ablation study of each component on SeqGeo and CrossRoad-66K datasets.}
\label{tab5}
\centering
\resizebox{0.98\linewidth}{!}{
\renewcommand{\arraystretch}{1.1}
\begin{tabular}{cccccccccccc}
\Xhline{1.5pt}
\multicolumn{3}{c}{Ablations}&
         \multicolumn{4}{c}{SeqGeo} & \multicolumn{4}{c}{CrossRoad-66K} \\ 
         \cline{4-12}
       Fusion & Aggregation & Position Encoding & Loss Function & R@1 & R@5 & R@10 & R@1\%   & R@1 & R@5 & R@10 & R@1\%     \\ 
\Xhline{0.8pt}               
        Concat&SFA & Compact&  Geo-InfoNCE	& 3.95\%&	12.46\%&	19.29\%&	50.86\%&	2.24\%&	7.27\%&	11.94\%&	29.35\%\\
        Random Mask&SFA& Sparse & Geo-InfoNCE	& 20.61\% &	40.33\%	&	49.21\%	&	76.52\%	&	5.01\%	&	14.28\%	&	21.12\%	&	39.42\%\\
        SCF&GAP & Compact& Geo-InfoNCE	& 3.55\%&	12.18\%&	18.54\%&	50.08\%&	3.11\%&	10.90\%&	16.97\%&	38.89\%   \\
        SCF&ViT-S & Compact & Geo-InfoNCE	& 12.24\%&	29.08\%&	38.79\%&	72.30\%&	3.18\%&	11.13\%&	16.81\%&	37.52\%   \\
        SCF&SFA & Compact &Soft-Margin Triplet	& 13.58\%&	29.80\%&	38.17\%&	70.37\%&	1.69\%&	5.56\%&	9.79\%&	25.40\%\\
        SCF&SFA & Compact & InfoNCE	& 17.70\%&	35.76\%&	44.32\%&	73.39\%&	2.99\%&	10.29\%&	16.43\%&	34.68\% \\
        SCF&SFA& Sparse & Geo-InfoNCE	& 20.89\% &	41.30\%	&	50.56\%	&	78.49\%	&	3.48\%	&	10.09\%	&	14.54\%	&	31.89\%\\
        SCF&SFA& Compact & Geo-InfoNCE	& \textbf{23.53\%} &	\textbf{44.30\%} &	\textbf{53.23\%} & \textbf{80.77\%} &	\textbf{6.20\%}  &	\textbf{16.83\%} &	\textbf{24.40\%} & \textbf{43.94\%}\\
\Xhline{1.5pt}
\end{tabular} 
}
\end{table*}

\subsection{Ablation Studies}\label{sec5_4}
In order to evaluate the effectiveness of each component in ASML, we performed extensive ablation studies. Moreover, considering the different numbers of ground images within a sequence in real-world scenarios, the recall rates of ASML on diverse ground image sequence lengths $K$ are shown in Table~\ref{tab6}.

\subsubsection{Ablation Study on Sequential Content Fusion Module (SCF)} In order to evaluate the effectiveness of our SCF module, we removed the SCF module and directly concatenated the local feature maps of the images in the sequence. As shown in Table~\ref{tab5}, the version integrated with the SCF module achieves gains of 19.58\% on the SeqGeo dataset and 3.96\% on the CrossRoad-66K dataset at R@1, respectively. The difference in performance gain between the two datasets may be related to their different scene distributions and visual characteristics. Compared with SeqGeo, CrossRoad-66K exhibits more complex scenes and more diverse ground-satellite appearance variations, which may affect the relative gain of the SCF module. This is consistent with the geo-localization results in Fig.~\ref{fig14}.

\begin{figure}[!h]
\centering
\includegraphics[width=0.98\linewidth]{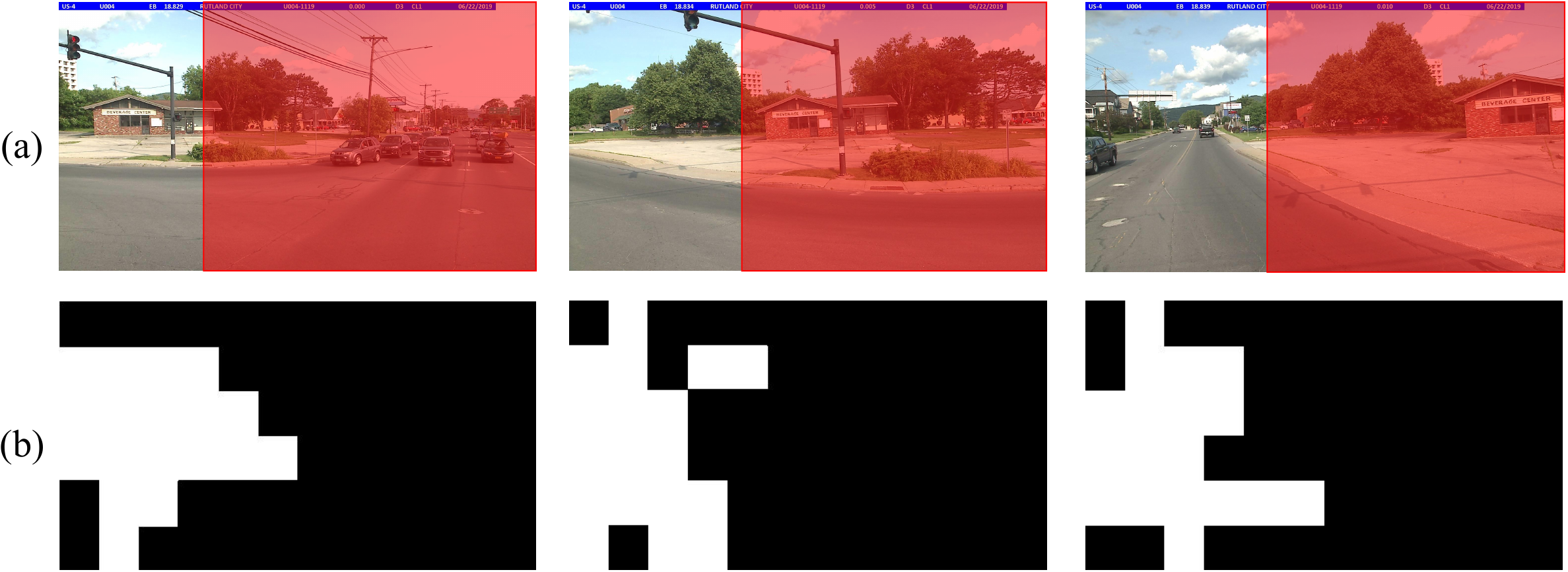}
\caption{Visualization of the attention masks on SeqGeo dataset. (a) shows the ground images. (b) shows the corresponding attention masks.}
\label{fig11}
\end{figure}

Fig.~\ref{fig11} visualizes representative feature-selection masks generated by SCF on the SeqGeo dataset. As illustrated in the figure, SCF tends to retain regions with relatively low cross-image correlation, which frequently correspond to complementary content that is less repeatedly observed in adjacent ground images. We note that such low-similarity regions are treated as a heuristic indicator of potentially complementary information rather than an exact estimate of geometric non-overlap.
\begin{table}[t]
\centering
\caption{Quantitative comparison of different feature-selection strategies on SeqGeo and CrossRoad-66K.}
\label{tab:scf}
\resizebox{0.99\linewidth}{!}{\begin{tabular}{cccccc}
\Xhline{1.3pt}
Dataset & Selection Strategy
& R@1 & R@5 & R@10 & R@1\% \\
\Xhline{0.8pt}

\multirow{3}{*}{SeqGeo}
& Random Mask
& 20.61\% & 40.33\%  & 49.21\%  & 76.52\%  \\
& High-Similarity SCF
& 17.07\%  & 35.85\%  & 44.87\%  & 72.77\%  \\
& \textbf{SCF}
& \textbf{23.53\%} &	\textbf{44.30\%} &	\textbf{53.23\%} & \textbf{80.77\%} \\

\Xhline{0.8pt}

\multirow{3}{*}{CrossRoad-66K}
& Random Mask
& 5.01\%  & 14.28\%  & 21.12\%  & 39.42\%  \\
& High-Similarity SCF
& 4.55\%  & 14.99\%  & 22.75\%  & 41.41\%  \\
& \textbf{SCF}
& \textbf{6.20\%} & \textbf{16.83\%} & \textbf{24.40\%} & \textbf{43.94\%} \\

\Xhline{1.3pt}
\end{tabular}}
\end{table}

To further verify whether the proposed feature-selection strategy contributes to geo-localization performance, we conduct a controlled comparison among SCF, random mask, and high-similarity SCF. All three variants retain the same number of feature patches and use identical subsequent feature reconstruction, aggregation, training, and evaluation settings, with only the feature-selection strategy being changed.

As shown in Table~\ref{tab:scf}, compared with the version equipped with random mask, SCF improves R@1 by 2.92\% and 1.19\% on SeqGeo and CrossRoad-66K, respectively, indicating that the performance gain cannot be attributed merely to discarding a subset of sequence features. Moreover, compared with high-similarity SCF, SCF improves R@1 by 6.46\% and 1.65\%, respectively. It demonstrates that suppressing highly correlated content and retaining lower-similarity, potentially complementary features, is more beneficial for sequence geo-localization. Therefore, the qualitative visualization and quantitative ablation jointly demonstrate the task-level effectiveness of the proposed feature selection strategy.

\subsubsection{Ablation Study on Spatial Feature Aggregation Module (SFA)} In order to evaluate the effectiveness of the SFA module, we first performed experiments by replacing it with a global average pooling layer (GAP) followed by a fully connected classifier. Then, we conducted experiments by replacing it with a ViT-S that does not contain the distillation token. As shown in Table~\ref{tab5}, compared to the version integrated with GAP, ASML achieves gains of 19.98\% on the SeqGeo dataset and 3.09\% on the CrossRoad-66K dataset at R@1. Compared to the version integrated with ViT-S, ASML achieves gains of 11.29\% on the SeqGeo dataset and 3.02\% on CrossRoad-66K dataset at R@1. These substantial improvements are mainly derived from that the SFA module enables adaptive cross-region information interaction among the global ground feature map, thus obtaining more discriminative feature descriptors. We also implemented experiments on the version equipped with sparse position encoding, \textit{i.e.,} the global feature map is encoded based on the order of the initial local feature maps, preserving their original positional relationships. It can be observed that the version with sparse position encoding performs worse than the compact position encoding. This is because the sparse position encoding introduces noisy or inconsistent location information into the attention aggregation process and weakens the discriminative ability of the final descriptor.


\begin{table}
\caption{Ablation study on different numbers $K$ of ground images in a sequence.}
\label{tab6}
\centering
\resizebox{0.88\linewidth}{!}{
\begin{tabular}{ccccccc}
\Xhline{1.5pt}
\multirow{2}{*}{$K$}&
         \multicolumn{3}{c}{SeqGeo} & \multicolumn{3}{c}{CrossRoad-66K} \\ 
         \cline{2-7}
       ~ & R@1 & R@5 & R@1\%   & R@1 & R@5 & R@1\%     \\ 
\Xhline{0.8pt}               
        2	& 12.45\%&	30.42\%&	69.45\%&	3.41\%&	10.73\%&	33.12\%\\
        3	& 15.86\%&	34.52\%&	74.07\%&	5.26\%&	14.61\%&	38.47\%  \\
        4	& 19.11\%&	37.55\%&	76.29\%&	\textbf{6.20\%}  &	\textbf{16.83\%}  & \textbf{43.94\%}\\
        5	& \textbf{23.53\%} &	\textbf{44.30\%} &	\textbf{80.77\%}&	4.21\%&	12.48\%&	38.94\% \\
        6	& 20.49\%&	39.99\%&	77.62\%&	3.67\%&	11.23\%&	35.95\%\\
\Xhline{1.5pt}
\end{tabular} 
}
\end{table}
\begin{table}
\caption{Comparison with existing methods on parameters, computational cost, and R@1 (on CrossRoad-66K dataset).}
\label{tab7}
\centering
\resizebox{0.98\linewidth}{!}{
\begin{tabular}{ccccc}
\Xhline{1.3pt}
Methods & Architecture & Parameters & GFLOPs& R@1\\
\Xhline{0.75pt}
         L2LTR\cite{ref27} & CNN+Transformer & 195.9M &	44.06   & 2.53\%  \\
         TransGeo\cite{ref28} & Transformer & \textbf{44.9M} &	11.34   & 2.37\% \\
         Sample4Geo\cite{ref29} & Transformer & 87.6M & 20.05   & 2.35\%  \\
         SeqGeo\cite{ref24} & CNN+Transformer &	47.02M &	10.30   &	1.86\%\\
         GeoDTR\cite{ref36} & CNN+Transformer &	48.5M &	38.89   &	3.15\%\\
         FlexGeo\cite{ref50} & Transformer &	87.6M &	140.14   &	3.82\%\\
         \textbf{ASML(Ours)} & CNN+Transformer & 54.2M &	\textbf{10.22} & \textbf{6.20\%}\\
\Xhline{1.3pt}
\end{tabular} 
}
\end{table}
\subsubsection{Ablation Study on Geo-InfoNCE Loss} In order to evaluate the effectiveness of Geo-InfoNCE loss, we compared it with the soft-margin triplet loss\cite{ref12} and the original InfoNCE loss, which are widely used in cross-view image geo-localization tasks. Quantitative comparisons are shown in Table~\ref{tab5}. Compared to the version trained with soft-margin triplet loss, ASML achieves gains of 9.95\%  on the SeqGeo dataset and 4.51\% on the CrossRoad-66K dataset at R@1. Compared to the version trained with the original InfoNCE loss, ASML achieves gains of 5.83\%  on the SeqGeo dataset and 3.21\% on the CrossRoad-66K dataset at R@1, respectively. The results indicate that the Geo-InfoNCE loss is superior to the other two losses, thus demonstrating the feasibility of the mixed-space alignment mechanism.

We further investigate the influence of the geographic term in Geo-InfoNCE. Since the feature distance term \(D_{i,j}\) and the geographic term \(G_{i,j}\) originate from different metric spaces, the coefficient \(\lambda_g\) explicitly controls the contribution of geographic term to the mixed-space metric. A small \(\lambda_g\) weakens the geographic constraint and makes the objective closer to conventional feature-based contrastive learning, while an excessively large \(\lambda_g\) may overemphasize geographic relationships at the expense of visual discrimination. To quantitatively evaluate this trade-off, we conducted a sensitivity analysis on $\lambda_g$, as shown in Table~\ref{tab_lambda}. These results indicate that the contribution of geographic information needs to be appropriately balanced with visual feature discrimination rather than simply strengthened. We select $\lambda_g$=0.75 because it yields the highest R@1 on both datasets and achieves a favorable overall balance across the evaluated retrieval metrics.
\begin{table}
\caption{Sensitivity analysis of varying scaling coefficient $\lambda_g$.}
\label{tab_lambda}
\centering
\resizebox{0.98\linewidth}{!}{
\begin{tabular}{ccccccc}
\Xhline{1.5pt}
\multirow{2}{*}{$\lambda_g$}& \multicolumn{3}{c}{SeqGeo} & \multicolumn{3}{c}{CVZY} \\ 
         \cline{2-7}
       ~ & R@1 & R@5 & R@1\%   & R@1 & R@5 & R@1\%     \\
\Xhline{0.8pt}               
        0.25 & 18.11\% &	36.39\% & 	73.85\% & 4.50\% & 13.61\% & 40.86\\
        0.5	& 20.00\% & 39.60\% & 77.90\% & 5.47\% & 15.81\% & 44.65\% \\
        0.75 &\textbf{23.53\%} &	\textbf{44.30\%} &	\textbf{80.77\%} & \textbf{6.20\%} & \textbf{16.83\%} & \textbf{43.94\%} \\
        1 &  23.32\% &	44.04\% &	80.48\% & 6.11\% & 16.67\% & 43.92\% \\
        1.25 & 20.17\% &	39.47\% &	78.39\% & 5.75\% & 17.14\% & 44.27\%\\
\Xhline{1.5pt}
\end{tabular} 
}
\end{table}

\begin{figure}[!h]
\centering
\includegraphics[width=0.95\linewidth]{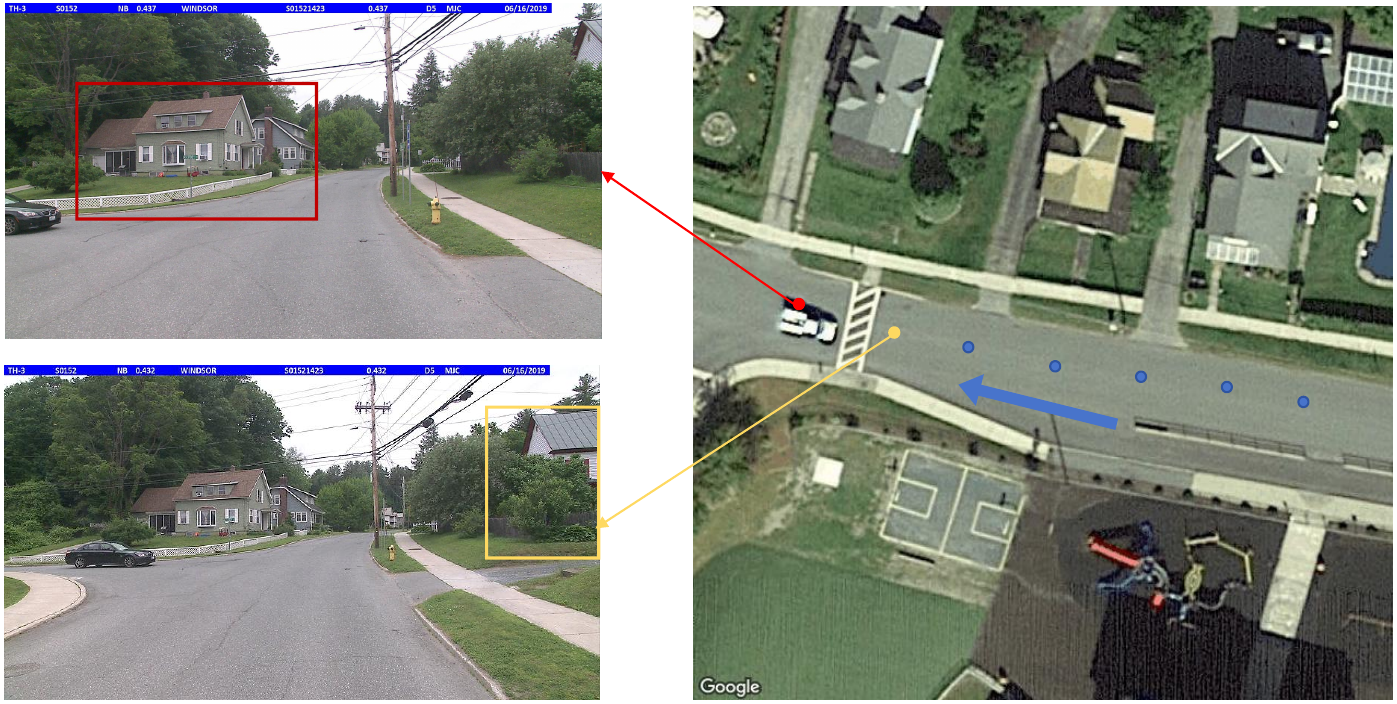}
\caption{Visualization of the last image in the ground image sequence and the satellite image. The red box in the left ground image represents irrelevant content between the ground image and the satellite image, while the points in the right satellite image represent the capture positions of each image in the ground image sequence.}
\label{fig13}
\end{figure}

\subsubsection{Ablation Study on the Sequence Length $K$} We performed experiments to test the impact of sequence length $K$ on geo-localization accuracy. Table~\ref{tab6} demonstrates the quantitative results on the SeqGeo and CrossRoad-66K datasets. It can be observed that ASML achieves the best geo-localization performance at $K=5$ on the SeqGeo dataset and $K=4$ on the CrossRoad-66K dataset. Fig.~\ref{fig13} reveals that the last two images capture many objects (\textit{e.g.}, vehicles, and buildings), which are not captured in the corresponding satellite images. Therefore, when $K$ is higher than the thresholds, it will introduce some redundant scenery features that do not exist in satellite images. In contrast, when $K$ is lower than the thresholds, the obtained ground image features cannot fully capture all the geographic-related features. In either case, it will reduce the correlation between ground image features and satellite image features, leading to a gradual decrease in geo-localization accuracy.

\subsection{Discussion}\label{sec5_6}
\subsubsection{Computational Efficiency}
In order to evaluate the computational efficiency of the proposed ASML, we compare the parameters, architecture, GFLOPs, and R@1 accuracy of ASML and several representative methods. As shown in Table~\ref{tab7}, the parameters of ASML are lower than L2LTR\cite{ref27}, Sample4Geo\cite{ref29}, and FlexGeo\cite{ref50}, and comparable to TransGeo\cite{ref28}, SeqGeo\cite{ref24}, and GeoDTR\cite{ref36}. However, because the proposed ASML uses smaller input image sizes, its computational costs are significantly lower than those of other methods, while the R@1 accuracy is also higher. This indicates that the proposed ASML has a higher computational efficiency and achieves higher geo-localization accuracy with lower computational costs.

\begin{table}[t]
\centering
\caption{Comparison of corner-road and straight-road scenarios under varying trajectory-change thresholds$\tau_{\theta}$.}
\label{tab8}
\resizebox{0.99\linewidth}{!}{
\begin{tabular}{llcccc}
\Xhline{1.3pt}
 Dataset &  Scenario &  Threshold$\tau_{\theta}$/Numbers 
&  R@1 &  R@5 &  R@1\% \\
\Xhline{0.8pt}  

\multirow{8}{*}{SeqGeo}
& \multirow{4}{*}{Corner}
&  $15^\circ$ (264) 
&  27.27\% &  46.21\% &  84.09\% \\
& & $20^\circ$ (106) 
& 17.93\% & 35.85\% & 84.91\% \\
& & $25^\circ$ (64) 
& 14.06\% & 31.25\% & 89.06\% \\
& & $30^\circ$ (41) 
& 17.07\% & 34.15\% & 90.24\% \\
\cline{2-6}
& \multirow{4}{*}{Straight}
&  $15^\circ$ (7,508) 
&  23.22\% &  43.33\% &  80.16\% \\
& & $20^\circ$ (7,666) 
& 23.42\% & 43.53\% & 80.22\% \\
& & $25^\circ$ (7,708) 
& 23.43\% & 43.65\% & 80.20\% \\
& & $30^\circ$ (7,731) 
& 23.39\% & 43.47\% & 80.24\% \\

\Xhline{0.8pt}  

\multirow{8}{*}{CrossRoad-66K}
& \multirow{4}{*}{Corner}
&  $15^\circ$ (479)
&  9.40\% &	25.47\% &	57.83\% \\
& & $20^\circ$ (342) 
& 7.90\% &	23.39\% &	56.43\% \\
& & $25^\circ$ (232) 
& 6.90\% &	23.28\% &	56.03\% \\
& & $30^\circ$ (155) 
& 6.45\% &	22.58\% &	56.13\% \\
\cline{2-6}
& \multirow{4}{*}{Straight}
&  $15^\circ$ (3,785) 
&  5.79\% &	15.67\% &	41.48\% \\
& & $20^\circ$ (3,922) 
& 6.03\% &	16.19\% &	42.17\% \\
& & $25^\circ$ (4,032) 
& 6.15\% &	16.39\% &	42.58\% \\
& & $30^\circ$ (4,109) 
& 6.18\% &	16.55\% &	42.83\% \\

\Xhline{1.3pt}
\end{tabular}
}
\end{table}
\begin{figure*}[t]
    \centering
    \subfloat[Non-overlap Precision (NoP)]{
        \label{fig12_1}
        \includegraphics[width=0.45\linewidth]{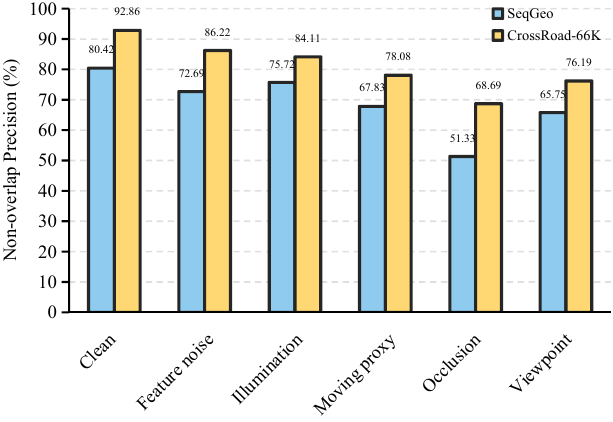}}
    \quad
    \subfloat[Overlap FSR (O-FSR)]{
        \centering
        \includegraphics[width=0.45\linewidth]{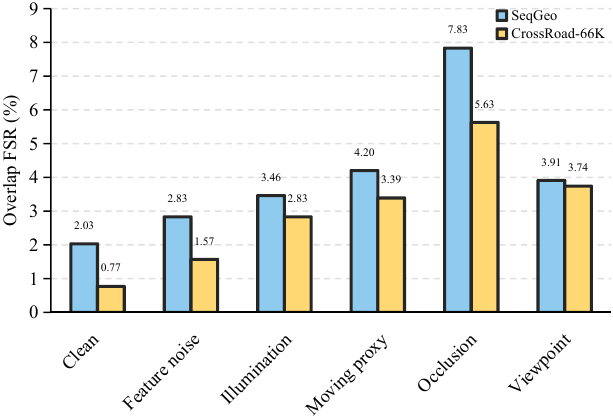}
        \label{fig12_2}}\\
    \subfloat[ROI $\Delta$FSR]{
        \centering
        \includegraphics[width=0.45\linewidth]{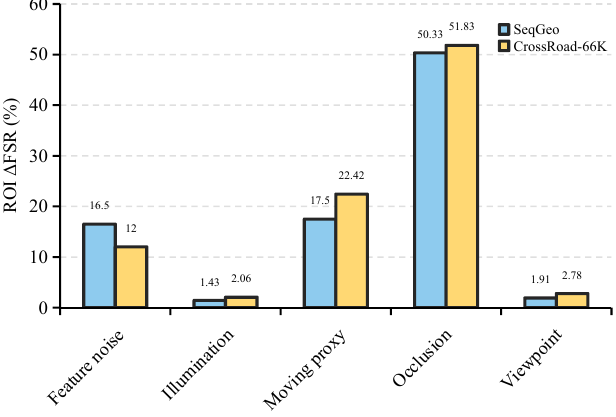}
        \label{fig12_3}}    
    \quad
    \subfloat[Aligned Mask IoU]{
        \centering
        \includegraphics[width=0.45\linewidth]{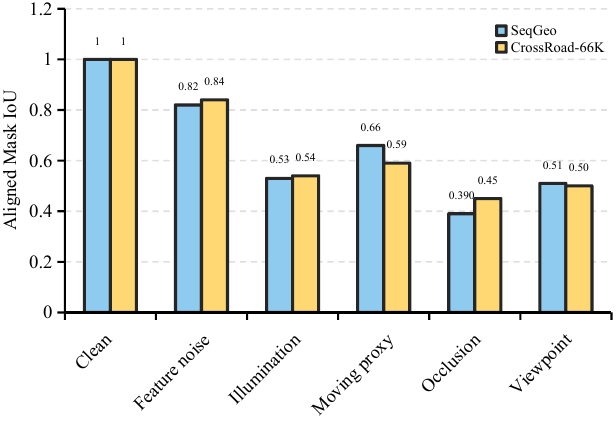}
        \label{fig12_4}}
    \caption{Quantitative analysis of SCF mask sensitivity to controlled perturbations on the SeqGeo and CrossRoad-66K datasets. (a) Non-overlap Precision (NoP) measures the proportion of selected features located in geometrically non-overlapping regions. (b) Overlap False Selection Rate (O-FSR) measures the proportion of overlapping features incorrectly selected. (c) ROI $\Delta$FSR measures the increase in false selection within the perturbed overlapping region relative to the clean input. (d) Aligned Mask IoU evaluates the spatial consistency between the clean and perturbed masks. Higher NoP and Mask IoU, and lower O-FSR and ROI $\Delta$FSR indicate better selection performance and robustness.}
    \label{fig12}
\end{figure*}

\subsubsection{Influence of Scenario}
As we mentioned in Sec.~\ref{sec3}, we investigate the influence of trajectory patterns on sequence geo-localization using four trajectory-change thresholds, \textit{i.e.,} $\tau_{\theta}\in\{15^\circ,20^\circ,25^\circ,30^\circ\}$. For each ground image sequence, we calculate its trajectory change $\Delta\theta$ and classify it as a corner-road sequence when $\Delta\theta\geq\tau_{\theta}$; otherwise, it is classified as a straight-road sequence. As shown in Table~\ref{tab8}, the number of corner-road samples decreases from 264 to 41 on SeqGeo and from 479 to 155 on CrossRoad-66K as the threshold increases. We report R@1, R@5, and R@1\% for each scenario using the same complete satellite gallery.

The results reveal distinct performance patterns across the two datasets. On CrossRoad-66K, corner-road sequences consistently outperform straight-road sequences across all three retrieval metrics under the evaluated thresholds, although the R@1 difference gradually narrows as the threshold increases. On SeqGeo, corner-road sequences achieve higher recall across all metrics at $15^\circ$. However, at stricter thresholds, straight-road sequences achieve higher R@1 and R@5, whereas corner-road sequences consistently maintain higher R@1\%. For example, at $30^\circ$, corner-road sequences achieve an R@1\% of 90.24\%, compared with 80.24\% for straight-road sequences.

These phenomena suggest that the influence of trajectory variation depends on both dataset SCENARIO. Turning trajectories can provide complementary observations that improve the coverage of correct locations within a broader retrieval shortlist, but this advantage does not necessarily translate into more accurate predictions. These results highlight the different contributions of trajectory patterns to candidate retrieval and fine-grained localization.

\begin{figure*}[t]
    \centering
    \subfloat[Query Image]{
        \label{fig_failure_1}
        \includegraphics[width=0.164\linewidth]{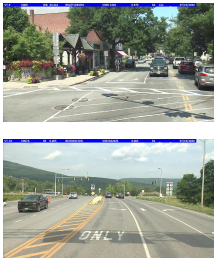}}
    \quad
    \subfloat[ASML]{
        \centering
        \includegraphics[width=0.3\linewidth]{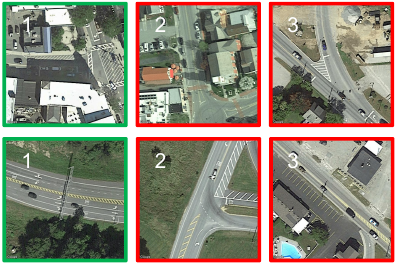}
        \label{fig_failure_2}}    
    \quad
    \subfloat[Removing SCF]{
        \centering
        \includegraphics[width=0.3\linewidth]{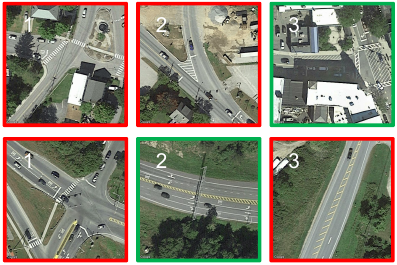}
        \label{fig_failure_3}}\\    
    \subfloat[Ground-Truth]{
        \centering
        \includegraphics[width=0.164\linewidth]{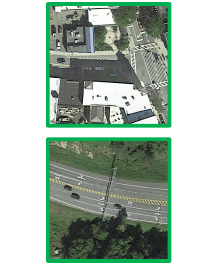}
        \label{fig_failure_4}}
    \quad
    \subfloat[Removing SFA]{
        \centering
        \includegraphics[width=0.3\linewidth]{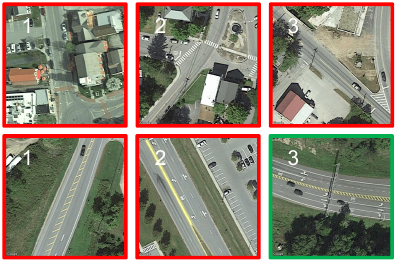}
        \label{fig_failure_5}}
    \quad
    \subfloat[Replacing Geo-InfoNCE with InfoNCE]{
        \centering
        \includegraphics[width=0.3 \linewidth]{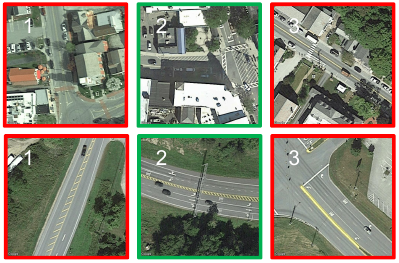}
        \label{fig_failure_6}}
    \caption{Visualization of module-specific failure cases. (a) shows the first images of the query ground image sequences. (b) shows the Top-3 results from ASML. (c) shows the Top-3 results from the version without SCF. (d) shows the ground-truth. (e) shows the Top-3 results from the version without SFA. (f) shows the Top-3 results from the version without Geo-InfoNCE. The green boxes indicate the ground-truth satellite images, while the red boxes denote incorrect retrieval results.}
    \label{fig:failure_case}
\end{figure*}

\subsubsection{Perturbation Robustness Analysis}
To investigate whether SCF confuses nuisance-induced feature dissimilarity with geometric novelty, we evaluate its robustness to five controlled perturbations: feature noise, illumination changes, synthetic occlusion, moving proxy (simulating local appearance changes through region replacement), and viewpoint changes (2-D perspective transformations). Each perturbation is applied only to the second ground image of each geometrically controlled image pair, while the first ground image remains unchanged. For each dataset, we sampled 150 held-out sequences and generated two geometrically controlled image pairs from one source ground image per sequence, using horizontal displacement ratios of 15\% and 25\%. This yielded 300 image pairs per dataset, with non-overlapping regions determined directly from the crop coordinates.

Then, we adopt four metrics to quantify the selection behavior: (1) \textbf{Non-Overlap Precision (NoP)}, measuring the proportion of selected features located in genuinely non-overlapping regions (higher is better); (2) \textbf{Overlap False Selection Rate (O-FSR)}, measuring the proportion of overlapping features incorrectly selected (lower is better); (3) \textbf{ROI $\Delta$FSR}, measuring the change in the selection rate within the perturbed overlapping region relative to the clean input (lower is better); and (4) \textbf{Aligned Mask IoU}, measuring the consistency between the clean and perturbed selection masks after spatial alignment (higher is better).

As shown in Fig.~\ref{fig12}, SCF exhibits a consistent preference for geometrically non-overlapping regions across both datasets. Under clean conditions, SCF achieves NoP values of 80.42\% and 92.86\% on SeqGeo and CrossRoad-66K, respectively, while maintaining low O-FSR values of 2.03\% and 0.77\%. Different perturbations affect the selection behavior to varying degrees. Feature noise has a relatively limited impact on mask consistency, with aligned mask IoU values of 0.82 and 0.84 on the two datasets. In contrast, illumination and viewpoint changes lead to more noticeable variations in the selected regions, although SCF still maintains a preference for non-overlapping content.

The moving proxy introduces localized selection errors, increasing ROI $\Delta$FSR by 17.50\% and 22.42\% on SeqGeo and CrossRoad-66K, respectively, while the overall O-FSR remains relatively low. This indicates that localized feature discrepancies can affect the selection mask without substantially changing the global selection statistics. Among all the perturbations, occlusion produces the most pronounced degradation. Compared with the clean condition, NoP decreases from 80.42\% to 51.33\% on SeqGeo and from 92.86\% to 68.69\% on CrossRoad-66K, accompanied by increases of 50.33\% and 51.83\% in ROI $\Delta$FSR, respectively. These results reveal a consistent selection pattern across the two datasets: SCF preferentially retains geometrically complementary features, while appearance-induced dissimilarity, particularly from occlusion, can also increase the selection of overlapping regions. This behavior reflects the inherent trade-off of similarity-based feature selection between suppressing redundant content and preserving complementary information under varying visual conditions. In summary, low cross-image similarity can indicate geometrically complementary content, but it may also arise from appearance variations or feature instability.

\subsubsection{Failure Case Analysis} 
To further investigate the behavior of each component, we conduct a module-specific failure analysis, as illustrated in Fig.~\ref{fig:failure_case}. Specifically, SCF and SFA are removed individually, while Geo-InfoNCE is replaced with InfoNCE and the other settings remain unchanged. For each query, we report the Top-3 retrieved satellite images obtained by the complete ASML and the corresponding ablated variants.

The failure cases demonstrate the complementary roles of SCF, SFA, and Geo-InfoNCE. In scenes dominated by repetitive roads, intersections, buildings, or vegetation, removing SCF increases confusion among visually similar candidates because redundant sequence observations are no longer explicitly suppressed. Removing SFA makes the model more dependent on local appearance similarity and less capable of distinguishing candidates with similar visual elements but different global spatial layouts. Replacing Geo-InfoNCE with InfoNCE, visually plausible hard negatives become more difficult to separate, showing the benefit of geographic supervision in constraining ambiguous feature-space matches.

These observations also demonstrate the limitations of each component. SCF may interpret low correlation caused by illumination changes, occlusions, moving objects, viewpoint variations, or feature instability as complementary information. SFA provides less benefit when spatial layouts are highly repetitive or informative features have already been discarded, while Geo-InfoNCE may be less effective for geographically close samples or noisy geographic annotations. In summary, the failure case analysis is consistent with the quantitative ablation results. SCF mainly reduces redundancy across sequence observations, SFA preserves discriminative spatial organization, and Geo-InfoNCE alleviates ambiguity through geographic supervision.


\section{Conclusion}\label{sec6}
In this work, we propose a novel cross-view image sequence geo-localization method ASML, which addresses the difficulties of capturing spatial layout information and eliminating redundant content interference. Unlike existing methods that overlook spatial coherence and simply concatenate sequence features, ASML utilizes the spatial attention correlations between the adjacent images in the image sequence to eliminate redundant content by the sequential content fusion module. Moreover, a Geo-InfoNCE loss function is designed to better guide the optimization of the model by simultaneously constraining feature similarity and geographic distance. We also establish a new dataset CrossRoad-66K to evaluate the performance of cross-view image sequence methods in complex scenarios. Extensive experiments demonstrate that ASML outperforms existing methods on both SeqGeo and CrossRoad-66K datasets. Ablation studies and controlled geometric evaluations further confirm the contributions of the proposed components and the preference of SCF for geometrically complementary content. Future work will investigate the integration of motion and visibility cues into adaptive feature selection to improve robustness in complex real-world environments.



\bibliographystyle{ieeetr}
\bibliography{reference}

\vfill

\end{document}